%% file: main.tex
\PassOptionsToPackage{table}{xcolor} %
\documentclass[sigconf, nonacm]{acmart}

\usepackage{caption} %
\usepackage{adjustbox}
\usepackage{graphicx}
\usepackage{caption}        %
\usepackage{adjustbox}      %
\usepackage[table]{xcolor}
\usepackage{tabularx}
\usepackage{makecell}
\usepackage{booktabs}
\usepackage{array}
\definecolor{tableheadcolor}{RGB}{189,215,238}
\usepackage{soul}
\usepackage{xcolor}
\sethlcolor{pink} %
\definecolor{tableheadcolor}{HTML}{B7D4EA} %
\rowcolors{2}{gray!10}{white}
\colorlet{tablealtcolor}{gray!10}

\definecolor{algored}{rgb}{1.0, 0.90, 0.90}
\definecolor{algoblue}{rgb}{0.85, 1.00, 1.00}
\definecolor{algogreen}{rgb}{0.875, 1.00, 0.875}  %

\newcommand{\hlword}[2]{%
  \begingroup
  \setlength{\fboxsep}{1pt}%
  \colorbox{#1}{\strut #2}%
  \endgroup
}

\input{bibsetup}
\input{copyright}

\input{packages}

\begin{document}
\title{Performance Analysis and Optimization of 3D Generative Diffusion Models across GPU Architectures}

\input{author}
\input{01.abstract}
\input{CCSXMLandKW}

\maketitle

\input{02.introduction}

\input{03.background}

\input{04.methodology}

\input{05.motivation}

\input{06.optimization}

\input{07.evaluation}

\input{08.related_work}

\input{09.conclusion}

\bibliographystyle{ACM-Reference-Format}
\balance
\bibliography{reference}

\end{document}

%% file: bibsetup.tex
\AtBeginDocument{%
  }

%% file: copyright.tex
\setcopyright{none}

%% file: packages.tex
\usepackage[final,commandnameprefix=ifneeded]{changes}

\usepackage{caption} %
\usepackage{adjustbox}
\usepackage{graphicx}
\usepackage{caption}        %
\usepackage{adjustbox}      %
\usepackage[table]{xcolor}
\usepackage{tabularx}
\usepackage{makecell}
\usepackage{booktabs}
\usepackage{array}
\definecolor{tableheadcolor}{RGB}{189,215,238}
\usepackage{soul}
\usepackage{xcolor}
\sethlcolor{pink} %
\definecolor{tableheadcolor}{HTML}{B7D4EA} %
\rowcolors{2}{gray!10}{white}
\colorlet{tablealtcolor}{gray!10}

\usepackage{graphicx} %
\usepackage[ruled, linesnumbered]{algorithm2e}
\usepackage{tikz}
\usetikzlibrary{shapes.geometric, arrows}
\usepackage{tcolorbox}
\usepackage{xcolor}

\definecolor{lpurple}{RGB}{235,225,255}

\usepackage{tikz}
\newcommand{\blackcircle}[1]{%
    \tikz[baseline=(char.base)]{
        \node[shape=circle,draw,fill=black,text=white,inner sep=1pt] (char) {#1};
    }%
}

\usepackage{booktabs}
\usepackage[table]{xcolor} 
\definecolor{lightblue}{RGB}{210,225,255}
\usepackage{graphicx}
\usepackage{makecell}
\usepackage{array}

%% file: author.tex
\author{Jeeho Ryoo}
\orcid{0009-0003-0401-3685}
\affiliation{%
  \institution{Fairleigh Dickinson University}
  \city{Vancouver}
  \state{BC}
  \country{Canada}
}
\email{j.ryoo@fdu.edu}

\author{Yongchan Jung}
\orcid{0009-0002-6504-6723}
\affiliation{%
  \institution{Fairleigh Dickinson University}
  \city{Vancouver}
  \state{BC}
  \country{Canada}
}
\email{y.jung@student.fdu.edu}

\author{Muhammad Ali Khaliq}
\orcid{0009-0006-2256-7974}
\affiliation{%
  \institution{The University of Colorado at Colorado Springs}
  \city{Colorado Springs}
  \state{CO}
  \country{USA}
}
\email{mkhaliq@uccs.edu}

\author{Weidong Zhang}
\orcid{0009-0009-2158-1784}
\affiliation{%
  \institution{Northeastern University}  
  \city{Vancouver}
  \state{BC}
  \country{Canada}
}
\email{zhang.weid@northeastern.edu}

\author{Jiatong Han}
\orcid{0009-0001-2843-1780}
\affiliation{%
  \institution{Fairleigh Dickinson University}
  \city{Vancouver}
  \state{BC}
  \country{Canada}
}
\email{j.han2@student.fdu.edu}

\author{Byeong Kil Lee}
\orcid{0000-0002-0260-2238}
\affiliation{%
  \institution{The University of Colorado at Colorado Springs}  
  \city{Colorado Springs}
  \state{CO}
  \country{USA}
}
\email{blee@uccs.edu}

\renewcommand{\shortauthors}{Jeeho Ryoo et al.}

%% file: 01.abstract.tex
\begin{abstract}
Diffusion models have become essential for high-fidelity 3D MRI synthesis, yet their deployment remains constrained by substantial GPU resource demands arising from hundreds of U-Net evaluations per sample and a highly heterogeneous kernel behavior. This paper performs a comprehensive performance analysis of the state-of-the-art medical diffusion model, Med-DDPM, across three generations of NVIDIA architectures to study kernel-level runtime breakdowns, instruction-mix characteristics, memory system utilization, warp-level activities, and profiler priority-score estimates. We show that training is overwhelmingly dominated by cuDNN convolution and implicit-GEMM kernels, with inefficiencies arising from memory-access patterns, tensor-layout conversions, and limited Tensor Core utilization. 
Guided by these insights, we evaluate two architecture-aware optimizations TF32 
Tensor Core activation and a 3D channels-last layout and demonstrate that they 
reduce SM cycles by up to 100$\times$, cut dynamic instructions by 100$\times$, 
raise Tensor Core utilization from 1.45 to 9.98$\times$, and increase IPC by 7\% on A100, all without degrading synthesis quality.
\end{abstract}

%% file: CCSXMLandKW.tex
\ccsdesc[500]{Computer systems organization~Parallel architectures}
\ccsdesc[500]{Computing methodologies~Machine learning}

\keywords{Performance Analysis; GPU Optimization; Tensor Cores; TF32; channels-last; Diffusion Models}

%% file: 02.introduction.tex
\begin{figure}[t]
  \centering
  \adjustbox{trim=2.5cm 12cm 2.5cm 2cm,clip,
             max width=\columnwidth,max height=0.23\textheight,center}{%
    \includegraphics{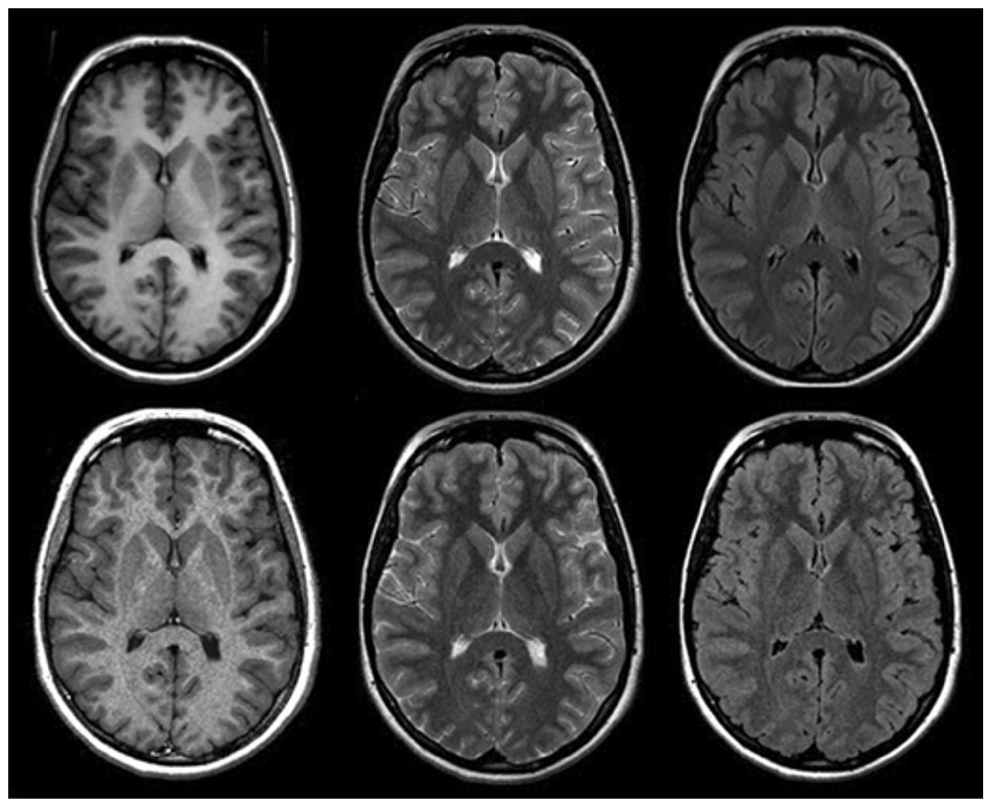}%
  }
  \caption{Conventional (top) and Mask-conditioned Synthetic (bottom) T1, T2, and FLAIR MRIs}
  \label{fig:mri_intro}
  \vspace{-5mm}
\end{figure}

\vspace{-2mm}
\section{Introduction}
Magnetic Resonance Imaging (MRI) is central to clinical neurology and neuroscience research, providing high-resolution 3D views of brain anatomy across complementary modalities such as T1-weighted, T2-weighted, and FLAIR scans~\cite{dorjsembe2024conditional}. These contrasts reveal tissue boundaries, fluid characteristics, and lesion patterns essential for diagnosis and longitudinal monitoring. However, large-scale MRI datasets remain difficult to assemble due to privacy constraints, scanner variability, long acquisition times, and the need for expert annotation~\cite{chung2023mri, qian2023physics}. To address these limitations, generative models have become an important tool for augmenting datasets with privacy-preserving images that reflect modality-specific appearance patterns. Figure~\ref{fig:mri_intro} illustrates these modality-specific contrasts by showing conventional T1, T2, and FLAIR slices alongside the corresponding mask-conditioned synthetic outputs.~\cite{chen2020augmentation, tmi2024specialissue}.

High-fidelity synthetic MRIs match anatomical boundaries, relative intensity profiles, and lesion-like hyperintensities, although they may differ subtly in texture or global scaling. These nuances motivate the evaluation through both quantitative metrics and qualitative expert review, as well as the use of explicit mask conditioning to control lesion placement and morphology for augmentation, harmonization, and privacy-preserving dataset expansion~\cite{yi2019gan_review}. Recent advances in denoising diffusion probabilistic models (DDPMs) have further improved the realism of 2D and 3D MRI synthesis, consistently demonstrating higher structural fidelity than GAN-based approaches~\cite{jiang2025frequencyaware}. Within this landscape, Med-DDPM stands out as a state-of-the-art yet still relatively new framework for 3D, mask-guided MRI synthesis, offering strong reconstruction quality and anatomically coherent outputs.

However, these accuracy benefits come with substantial computational cost. A single diffusion sample requires hundreds of sequential U-Net evaluations, each executing dozens of 3D convolutions, normalization layers, and elementwise operations~\cite{webber2024dmrecon}. This results in heterogeneous GPU execution patterns with frequent kernel launches, alternating compute-bound and memory-bound phases, high arithmetic intensity for convolutions, and bandwidth-limited behavior for normalization and reduction kernels. Consequently, training 3D diffusion models demands hundreds of GPU-hours even for modest datasets. Model improvements alone cannot resolve these bottlenecks as their practical deployment depends critically on architectural features such as tensor-core design, memory bandwidth, layout handling, and precision modes across GPU generations~\cite{hanindhito2024tensorcores}.

This systems perspective is largely absent from existing medical imaging research, where diffusion models are typically evaluated only on image quality. In contrast, this work performs a full-stack GPU performance study of Med-DDPM across three NVIDIA architectures Volta (V100), Ampere (A100), and Hopper (H100) using NVIDIA's Nsight Compute profiler to characterize the model from kernel level down to microarchitectural scheduling behavior. Beyond kernel-level execution characteristics, we analyze detailed kernel mixes, dynamic instruction-mix profiles, priority-score distributions, SM-level utilization, memory-hierarchy behavior, warp occupancy and warp-scheduler efficiency, and stall-type breakdowns involving long-latency dependencies, math-pipeline throttling, and divergent-branch resolution. These multi-level measurements reveal that Med-DDPM is overwhelmingly dominated by cuDNN 3D convolution and implicit-GEMM kernels, and that inefficiencies arise not only from memory-access irregularities, tensor-layout conversions, and precision-mode constraints, but also from fragmented warp scheduling, cache residency effects, and underutilized Tensor Core pipelines on Ampere/Hopper architectures. Building on these profiling results, we apply two targeted, architecture-aware interventions that directly address the dominant bottlenecks identified in our analysis. We show how enabling Tensor Core execution on Ampere/Hopper and restructuring Med-DDPM’s memory layout fundamentally changes the balance of compute-, memory-, and scheduler-driven behavior across the model. These findings lead to the following contributions:

\begin{itemize}
    \item We present a detailed system-level and microarchitectural GPU characterization of Med-DDPM across three different architectures (V100, A100, and H100).
    \item We identify architecture-specific bottlenecks in convolution, normalization, and layout-conversion kernels using kernel-mix breakdowns, IPC stacks, and priority-score analysis.
    \item We apply and evaluate two system level optimizations TF32 Tensor Core activation and a 3D channels-last layout derived from microarchitectural implications.
    \item We show that TF32 Tensor Core activation reduces SM cycles by up to \(\sim5\times\) on A100, while a 3D channels-last layout exposes new memory-bound microkernels that require further fusion to reach peak throughput.
    \item We analyze architectural behavior shifts with the proposed optimizations in resource utilization, kernel mix, memory, active warps, and stalls. Based on the analysis, practical guidelines are presented for optimizing general 3D diffusion models on modern GPUs.
    
\end{itemize}

%% file: 03.background.tex
\section{Background}
\subsection{Med-DDPM Architecture}
Diffusion models have emerged as a leading paradigm for medical image synthesis due to their stability and ability to preserve anatomical structure, outperforming GAN-based approaches that often exhibit mode collapse and training instability in clinical settings ~\cite{ho2020ddpm, nichol2021improved, jiang2021quantization}. These limitations motivate the shift toward diffusion-based approaches, which provide more stable optimization behavior and stronger structural fidelity. Med-DDPM adapts this paradigm for 3D brain MRI generation under semantic guidance, using segmentation masks to address challenges such as limited annotated datasets and privacy concerns while enabling explicit control over structural placement ~\cite{dorjsembe2024conditional, ronneberger2015unet}. Beyond sampling stability, diffusion models can integrate structured guidance signals that further improve anatomical control. This mask-conditioned generation strategy allows the model to synthesize anatomically faithful and pathology-aware MRIs suitable for augmentation, controlled simulation, and dataset expansion in clinical workflows ~\cite{cicek20163dunet}.

\begin{figure}[tbp]
    \centering
    \includegraphics[width=1\linewidth,  keepaspectratio]{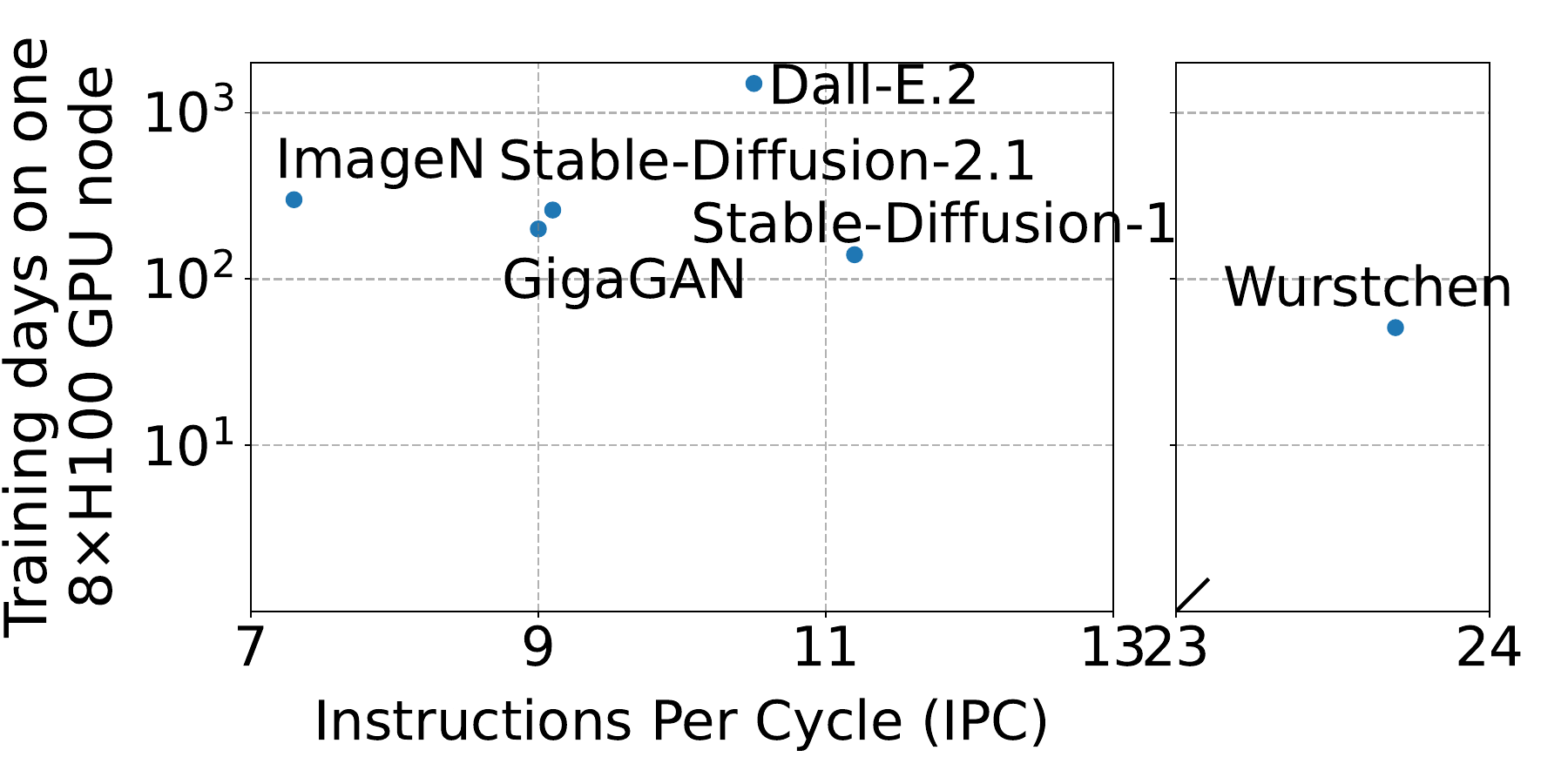}
    \vspace{-20pt}
    \caption{
        IPC versus Training Duration for Representative Generative Models
        }
    \label{fig:fid_vs_training}
\end{figure}

Diffusion models are highly compute-intensive: each synthesized sample requires hundreds to thousands of iterative denoising steps, so every training batch executes repeated full U-Net passes rather than the single forward–backward cycle typical of GANs. For 3D MRI, this cost grows further as large volumetric tensors stress memory bandwidth, kernel-launch rates, and SM utilization. Figure~\ref{fig:fid_vs_training} situates this burden by comparing training duration and IPC across diffusion and non-diffusion generative models~\cite{Sehwag2025MicroBudget}, revealing that diffusion architectures cluster in a medium-IPC, long-duration regime.

Med-DDPM inherits this algorithmic structure, motivating the detailed GPU-level analysis in this paper. Even well-optimized diffusion models require days of uninterrupted GPU time and exhibit wide hardware-efficiency variation. In contrast, models like Wurstchen achieve higher IPC and much shorter training times because they avoid iterative denoising. This fundamental disparity iterative refinement versus single-pass generation explains why diffusion models are slower, less compute-efficient, and far more resource-intensive despite their superior fidelity, posing a major scalability challenge for large 3D medical generators such as Med-DDPM~\cite{muellerfranzes2023comparison}.

Architecturally, Med-DDPM follows a standard DDPM forward process with a cosine noise schedule, gradually corrupting a clean MRI volume with Gaussian noise before reversing the corruption through a conditioned 3D U-Net ~\cite{dorjsembe2024conditional}. The reverse network predicts the noise added at each timestep while concatenating a one-hot encoded segmentation mask with the noisy input, ensuring anatomical awareness throughout the denoising process ~\cite{yi2019gan_review}. This design enables Med-DDPM to generate structurally coherent MRIs even from limited training data, producing higher-quality images and stronger downstream segmentation performance than GAN-based and latent-diffusion baselines ~\cite{dorjsembe2024conditional}.

\begin{figure}[tbp]
  \centering
\includegraphics[width=\linewidth]{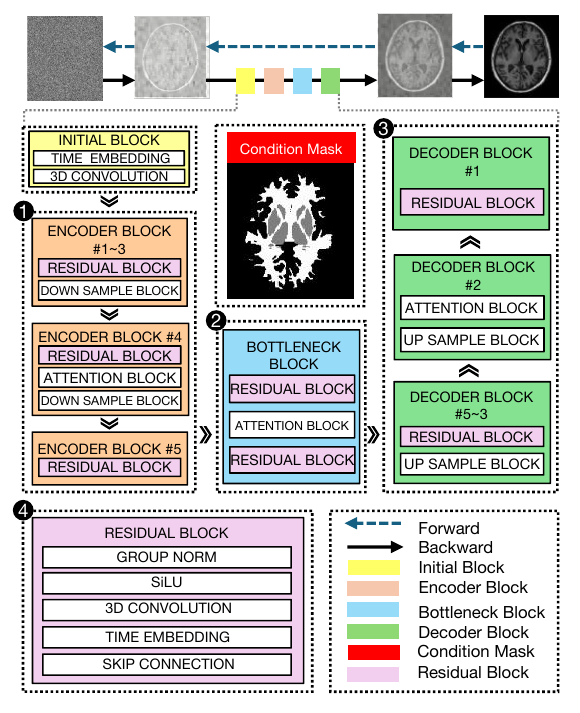}
  \vspace{-8mm}
  \caption{Overview of the Med-DDPM architecture}
  \Description{Diagram showing the Med-DDPM architecture with forward noise schedule, segmentation mask conditioning, and reverse denoising via 3D U-Net.}
  \label{fig:med_ddpm_arch}
  \vspace{-5mm}
\end{figure}

Figure~\ref{fig:med_ddpm_arch} summarizes the full workflow, pairing the forward diffusion trajectory with the conditioned reverse denoising pathway. In the reverse phase, the U-Net processes mask-aware feature maps through the Initial Block and Encoder Blocks \blackcircle{1}, composed of Residual Blocks \blackcircle{4}, Down-sample layers, and Attention Blocks, before integrating global features in the Bottleneck Block \blackcircle{2}. Decoder Blocks \blackcircle{3} then reconstruct the anatomical volume using skip connections, with the segmentation mask guiding feature refinement at each timestep ~\cite{wang2003msssim, goodfellow2014gan}. This conditioning mechanism enforces anatomical boundaries and pathological regions during sampling, enabling Med-DDPM to generate clinically realistic, mask-guided 3D MRI variations despite the substantial computational burden associated with diffusion training.

Unlike CNNs or transformers, diffusion models execute long chains of heterogeneous kernels, large convolutions, normalization layers, elementwise ops, and small reductions repeated hundreds of times per sample. This creates alternating compute-bound and memory-bound phases that expose architectural bottlenecks not visible in conventional training workloads. A detailed GPU-profiling study is therefore essential to understand how diffusion workloads interact with Tensor Cores, memory hierarchy, warp scheduling, and cache behavior.

\vspace{-3mm}
\subsection{GPU Architectures}
Diffusion models stress GPUs differently from conventional CNN or transformer workloads. Because Med-DDPM consists of hundreds of sequential U-Net evaluations, its computational footprint interacts strongly with the underlying GPU architecture. Each sampling trajectory executes hundreds of U-Net blocks composed of convolutions, GroupNorm, residual connections, and pointwise operations, producing a rapid stream of small and medium-sized kernels with alternating compute and memory-bound behavior. As a result, the achievable performance depends closely on core GPU architectural features such as SM organization, tensor-core capabilities, memory hierarchy, cache design, and supported precision modes. Table~\ref{tab:gpu-arch-compare} summarizes the architectural characteristics of the three NVIDIA GPU generations evaluated in this study~\cite{nvidia2023hopper}.

Modern NVIDIA GPUs organize computation around Streaming Multiprocessors (SMs), which integrate scalar and vector ALUs, warp schedulers, register files, shared memory, and L1 cache. The L1 cache size has increased across generations from 128KB per SM in Volta to 192KB in Ampere and 256KB in Hopper, improving intra-block data reuse and reducing latency for small kernel operations. These architectural factors directly influence how efficiently diffusion workloads can be scheduled and executed. Convolutions in U-Net blocks map to implicit GEMM operations, making tensor-core throughput a primary determinant of overall diffusion speed. Volta’s V100 provides Gen1 tensor cores optimized for FP32 and INT8, Ampere’s A100 introduces Gen3 tensor cores with TF32 and INT8 support, and Hopper’s H100 adds Gen4 tensor cores with FP8 and INT4 acceleration. These architectural differences directly influence how effectively diffusion models can exploit mixed-precision execution without degrading denoising stability~\cite{zhang2023precision}.

\begin{table}[t]
  \centering
  \footnotesize
  \setlength{\tabcolsep}{-1pt}        %
  \renewcommand{\arraystretch}{1.05} %
  \caption{Comparison of GPU Architectures and Features}
  \label{tab:gpu-arch-compare}
  \rowcolors{2}{tablealtcolor}{white} %
  \begin{tabular}{|>{\centering\arraybackslash}p{0.34\linewidth}|
                      >{\centering\arraybackslash}p{0.22\linewidth}|
                      >{\centering\arraybackslash}p{0.22\linewidth}|
                      >{\centering\arraybackslash}p{0.22\linewidth}|}
    \hline
    \rowcolor{tableheadcolor}
    \textbf{Feature} & \textbf{H100} & \textbf{A100} & \textbf{V100} \\
    \hline
    Architecture                  & Hopper             & Ampere              & Volta            \\
    \hline
    CUDA Cores                    & 14592              & 6912                & 5120             \\
    \hline
    Memory Size (GB)              & 80                 & 40 / 80             & 16 / 32          \\
    \hline
    Memory Type                   & HBM3               & HBM2e               & HBM2             \\
    \hline
    Memory BW (GB/s)              & 3,350              & 2,000               & 900              \\
    \hline
    Peak FP32 (TFLOPS)            & 60                 & 19.5                & 15.7             \\
    \hline
    Tensor Cores                  & Gen4 + FP8         & Gen3                & Gen1             \\
    \hline
    NVLink BW (GB/s)              & 900                & 600                 & 300              \\
    \hline
    PCIe Version                  & Gen5               & Gen4                & Gen3 / Gen4      \\
    \hline
    Multi-Instance GPU            & Yes                & Yes                 & No               \\
    \hline
    Transformer Engine            & Yes                & No                  & No               \\
    \hline
    L1 Cache (per SM)             & 256 KB             & 192 KB              & 128 KB           \\
    \hline
    L2 Cache                      & 50 MB              & 40 MB               & 6 MB             \\
    \hline
    \rowcolor{tablealtcolor}
    Supported Precision           & \makecell[c]{FP16/32/64, TF32\\INT8, INT4, FP8} 
                                  & \makecell[c]{FP16/32/64, TF32\\INT8, INT4} 
                                  & \makecell[c]{FP16/32/64      \\INT8} \\
    \hline
  \end{tabular}
  \vspace{-3mm}
\end{table}

The memory hierarchy is equally important because many kernels in diffusion models (elementwise ops, normalization layers, and layout conversions) are memory-bound. HBM bandwidth increases substantially across generations (V100’s 900~GB/s, A100’s 2,000~GB/s, H100’s 3,350~GB/s), and L2 cache capacity grows from 6~MB in Volta to 40~MB in Ampere and 50~MB in Hopper. These improvements reduce stalls for bandwidth-limited stages of the U-Net and improve feature-map reuse across layers~\cite{liu2021l1cache}.

Precision support also evolves across generations. V100 primarily targets FP32 and INT8 execution, A100 adds TF32 and INT4 paths for both training and inference, and H100 enables FP8 acceleration for tolerant layers. Diffusion models accumulate numerical error across timesteps, so the availability of multiple precision modes is critical for balancing stability and throughput in medical imaging tasks~\cite{jia2016memoryhierarchy}.

Finally, diffusion workloads generate a high frequency of small kernel launches. Higher SM counts and improved warp scheduling mechanisms in A100 and H100 help sustain throughput under such fragmented execution patterns. For 3D medical diffusion models like Med-DDPM where tensor volumes are large and precision demands strict, these architectural differences have direct impact on achievable sampling time and fidelity. Overall, the progression from Volta to Ampere to Hopper provides increasingly favorable conditions for high-resolution diffusion workloads~\cite{dally2023nvidia}.

%% file: 04.methodology.tex
\section{Experimental Methodology}
This section describes the experimental methodology used to profile Med‑DDPM training across H100, A100, and V100. The methodology covers hardware and software setup, measurement strategy, and profiling configuration.

\subsection{Hardware and Software Setup}
The profiling and performance evaluations were performed on three platforms summarized in Table~\ref{tab:gpu-arch-compare}. The profiling on H100 and A100 was performed inside Docker containers on Indiana Jetstream2 and TACC Stampede3, while the V100 runs used a Singularity/Apptainer image on SDSC Expanse to comply with site policies~\cite{nvidia2022h100, nvidia2020a100, nvidia2017v100}. Since Docker relies on a root-privileged daemon, many sites restrict its use due to security concerns. Instead, they support user-level containers such as Singularity/Apptainer, which align better with system security policies. We used those containerized environments to ensure that the Med-DDPM workload, software dependencies, and CUDA/cuDNN versions remained bit-identical across all GPU architectures, eliminating run-to-run variability caused by divergent system libraries or site-specific configurations. The software stack for all runs used PyTorch 2.0.1 and Python 3.11.4. Nsight Compute CLI (\texttt{ncu}) version 2025.2.1 (CUDA Toolkit 12.9, cuDNN version 8.9.2) was used for profiling in the kernel-level, architecture-level, and instruction-level~\cite{nvidia2025nsightguide}. For every kernel-level profile, Nsight Compute provides instruction mix, warp execution efficiency, SM utilization, Tensor core usage, etc. Microarchitecture-level profiling information includes pipeline stalls and L1/L2 cache behavior, while SASS (machine instructions) and per-instruction stall reasons are provided at instruction-level~\cite{nvidia2023metrics}.

\subsection{Performance Measurement}
Med-DDPM training was executed with the same hyperparameters across platforms: 25 epochs, \texttt{batchsize=1}, \texttt{input\_size=128}, \texttt{depth\_size=128}, \texttt{num\_channels=64}, and the same \texttt{resume\_weight}. The profiling script invoked Nsight Compute with the following key options: \texttt{--set full}, \texttt{--call-stack}, \texttt{--nvtx}, \texttt{--csv}, \texttt{--export}, \texttt{--launch-skip 10}, \texttt{--launch-count 100}, and \texttt{--target-processes all}. The option \texttt{--launch-skip 10} avoids warm-up noise, while \texttt{--launch-count 100} provides statistical stability for repeated kernel launches~\cite{yang2020rooflinecpe}. 
We could not collect data using such performance monitoring counters \texttt{smsp\_\_warps\_issue\_stalled\_long\_scoreboard.sum}, \texttt{smsp\_\_warps\_issue\_stalled\_math\_pipe\_throttle.sum}, and \texttt{smsp\_\_warps\_issue\_stalled\_branch\_resolving.sum} as they do not exist in a relatively old V100 architecture. Thus, we do not report these metrics in \S\ref{subsec:schedule_analysis}.

\begin{figure}[t]
  \vspace{-2mm}
  \centering
\includegraphics[width=\linewidth]{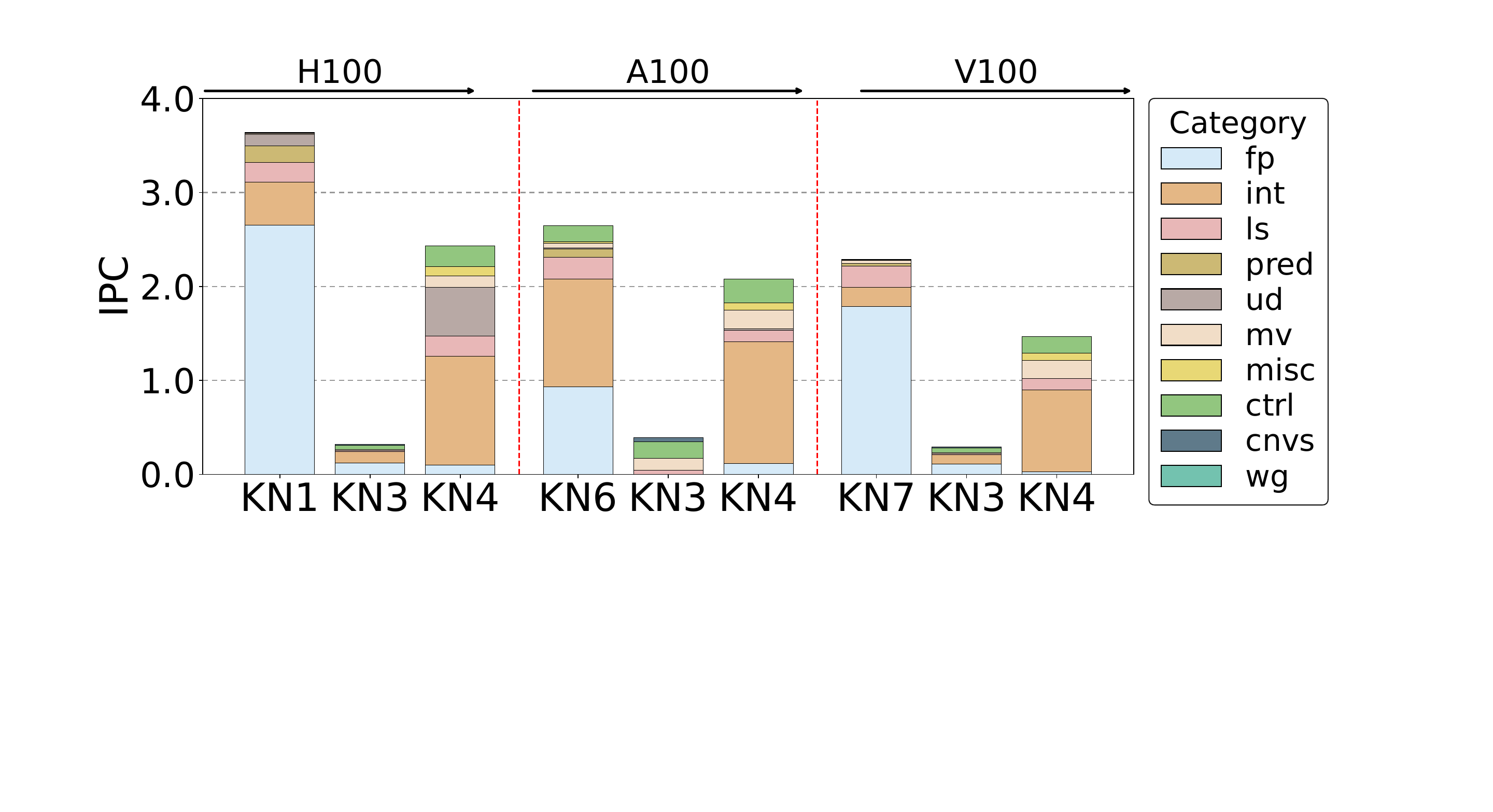}
  \vspace{-60pt}
  \caption{IPC Stack Bars}
  \vspace{-3mm}
  \label{fig:ipc_breakdown}
\end{figure}
The Med-DDPM training workload generates many short, recurrent kernels primarily convolutions, elementwise operations, and reductions making kernel-level profiling essential for understanding end-to-end behavior. For each GPU platform, we used Nsight Compute (\texttt{ncu}) under the same training configuration to collect per-kernel execution statistics and microarchitectural metrics. The primary metrics for cross-platform comparison were kernel duration, instruction count, and cycle count, which we used to compute per-kernel IPC (instructions per cycle) and an overall IPC via the geometric mean of per-kernel IPC. In addition, we recorded memory bandwidth by kernel duration, and the profiler’s priority score, which ranks kernels by estimated optimization potential (See \S\ref{subsec:sys_perf})~\cite{yang2020roofline}. Together, these measurements provide a consistent basis for characterizing kernel behavior and comparing training performance across architectures.

%% file: 05.motivation.tex
\vspace{-2mm}
\section{Performance Analysis of Med-DDPM}
\label{sec:bottleneck}
Diffusion-driven MRI synthesis places sustained pressure on GPUs because each reverse-diffusion step invokes a full 3D U-Net pass composed of convolutional blocks, normalization stages, and auxiliary tensor operations. Although modern samplers reduce the total number of denoising steps, the execution remains dominated by a large collection of kernels whose behavior varies significantly across architectures. To understand how these workloads interact with contemporary GPUs and where inefficiencies arise, we now shift from algorithmic context to a detailed systems-level study. This section presents a comprehensive performance analysis of Med-DDPM, examining kernel runtimes, instruction behavior, and priority-score profiles to reveal the architectural bottlenecks that govern end-to-end training throughput.

\begin{table}[t]
  \centering
  \footnotesize
  \caption{Kernel Name Abbreviations}
  \vspace{-2mm}
  \label{tab:kernel-abbrev}
  \rowcolors{2}{gray!10}{white}
  \begin{tabular}{|p{0.08\linewidth}|p{0.35\linewidth}|p{0.08\linewidth}|p{0.32\linewidth}|}
    \hline
    \rowcolor{tableheadcolor}
    \textbf{Abbr} & \textbf{Kernel Name} & \textbf{Abbr} & \textbf{Kernel Name} \\
    \hline
    KN1 & \texttt{sm80\_xmma\_fprop\_implicit} & KN5 & \texttt{nhwcToNchwKernel} \\
    \hline
    KN2 & \texttt{sm90\_xmma\_fprop\_implicit} & KN6 & \texttt{implicit\_convolveNd} \\
    \hline
    KN3 & \texttt{RowwiseMomentsCUDAKernel}         & KN7 & \texttt{\_5x\_cudnn\_volta\_scudnn} \\
    \hline
    KN4 & \texttt{elementwise\_kernel}              & KN8 &  \texttt{nhwcAddPaddingKernel } \\
    \hline
  \end{tabular}
  \vspace{-3mm}
\end{table}

\vspace{-2mm}
\subsection{System Performance Analysis}
\label{subsec:sys_perf}
We extracted instruction-per-cycle (IPC) stack profiles for representative kernels in Med-DDPM’s reverse-diffusion loop. Table~\ref{tab:kernel-abbrev} shows representative kernel name abbreviations used in the profiling. As illustrated in Figure~\ref{fig:ipc_breakdown}, diffusion workloads exhibit persistent inefficiencies across architectures. On the V100, the dominant cuDNN 3D convolution incurs \(\sim 3.9\times 10^{9}\) cycles with an aggregate IPC of 2.38, of which nearly 78\% are FP instructions indicating that the model saturates neither memory pipelines nor Tensor-Core pathways. On the A100, the corresponding tensor-core GEMM completes in \(\sim 1.18\times 10^{9}\) cycles (a \(3.3\times\) reduction), yet its IPC stack (INT: 42\%, FP: 25\%) shows that memory movement and reduction operations still occupy a significant fraction of execution. The H100 further reduces execution to \(\sim 8.2\times 10^{8}\) cycles, but its IPC decomposition shows more than 50\% of instructions in UD (tensor-core micro-ops) and only 2\% in classical FP pipelines, with the remainder dominated by LS, MV, and CTRL instructions. Across all GPUs, diffusion kernels achieve suboptimal utilization because memory access, layout conversions, and repeated small reductions impose structural bottlenecks that compute improvements alone cannot overcome.

\begin{figure}[t]
  \centering
  \includegraphics[width=\linewidth, keepaspectratio]{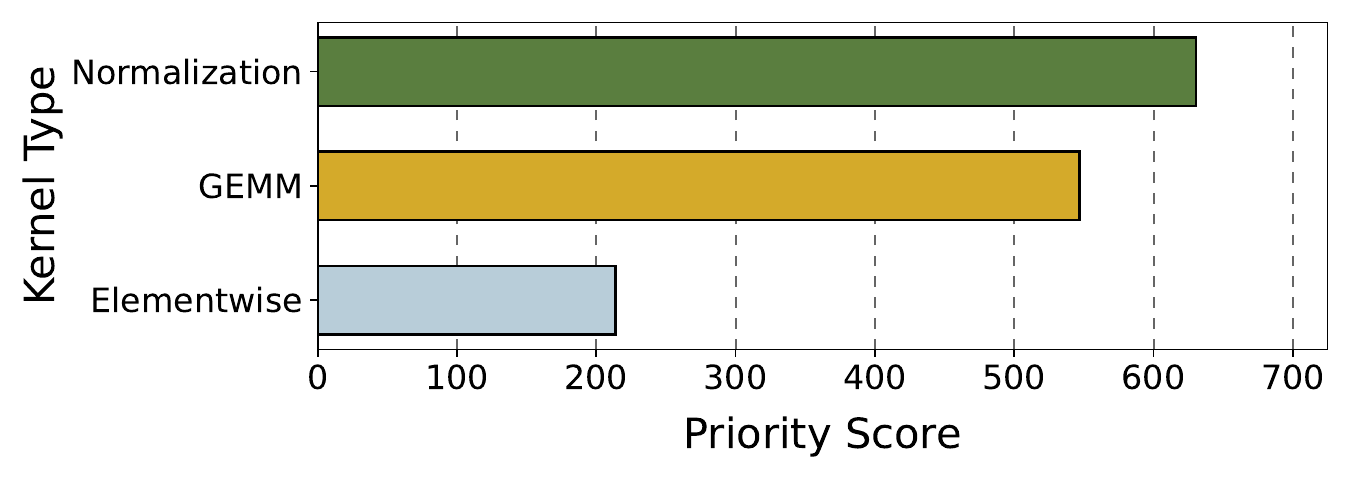}
\vspace{-20pt}
  \caption{Priority Scores of Med-DDPM}
  \vspace{-10pt}
  \label{fig:priority_score}
\end{figure}

To understand not only how kernels behave, but also which kernels matter most, we analyze Nsight Compute’s \textit{priority score}. This metric estimates potential optimization benefit by combining kernel duration and the percentage of achievable speedup as shown in Equation \ref{eq:priority_score}.
\begin{equation}
\label{eq:priority_score}
\footnotesize
\text{Priority Score} = \text{Duration} \times \text{Estimated Speedup}
\end{equation}
Nsight Compute computes the \textit{Estimated Speedup} by comparing each kernel’s measured throughput against the maximum throughput that the GPU could theoretically sustain for its dominant bottleneck. Formally, Equation \ref{eq:estimated_speedup} explains how the profiler quantifies this remaining performance headroom as clamped to the range [0,1] to represent the percentage of potential improvement. A kernel operating near its hardware limit yields an estimated speedup close to zero, whereas kernels far below their achievable ceilings exhibit larger values and therefore higher optimization potential. 
\begin{equation}
\label{eq:estimated_speedup}
\footnotesize
\text{Estimated Speedup} \approx
\frac{\text{Achievable Performance}}{\text{Observed Performance}} - 1
\end{equation}
Summing these scores across kernel categories allows us to quantify total optimization opportunity. The result for Med-DDPM is shown in Figure~\ref{fig:priority_score} where normalization kernels account for the largest cumulative priority, followed by tensor-core GEMMs and then elementwise operations. This reveals that diffusion workloads are not dominated by a single computational primitive, but rather by a heterogeneous mixture of memory-bound, reduction-bound, and shape-sensitive kernels.

Taken together, the IPC and priority-score patterns highlight that faster samplers or improved denoising schedules cannot, by themselves, resolve Med-DDPM’s inefficiencies. As long as the underlying kernels remain memory-bound, reduction-heavy, or dominated by non-compute micro-operations, algorithmic refinements will yield only incremental gains. Likewise, moving to newer GPUs such as A100 or H100 does not automatically eliminate bottlenecks when substantial fractions of instructions arise from layout handling, address generation, or control flow rather than arithmetic work. These observations frame the broader motivation for our study that synthetic MRI generation's diffusion-based pipelines remain prohibitively expensive to train at scale.

\noindent
\fcolorbox{black}{lpurple}{%
  \parbox{\dimexpr\linewidth-2\fboxsep-2\fboxrule\relax}{%
    \textbf{Takeaway 1:} Med-DDPM underutilizes GPU pipelines across all architectures due to persistent memory, layout, and reduction overheads.
    }%
}

\subsection{Kernel-Level Analysis}

\begin{figure}[t]
  \centering
  \includegraphics[width=0.8\linewidth, keepaspectratio]{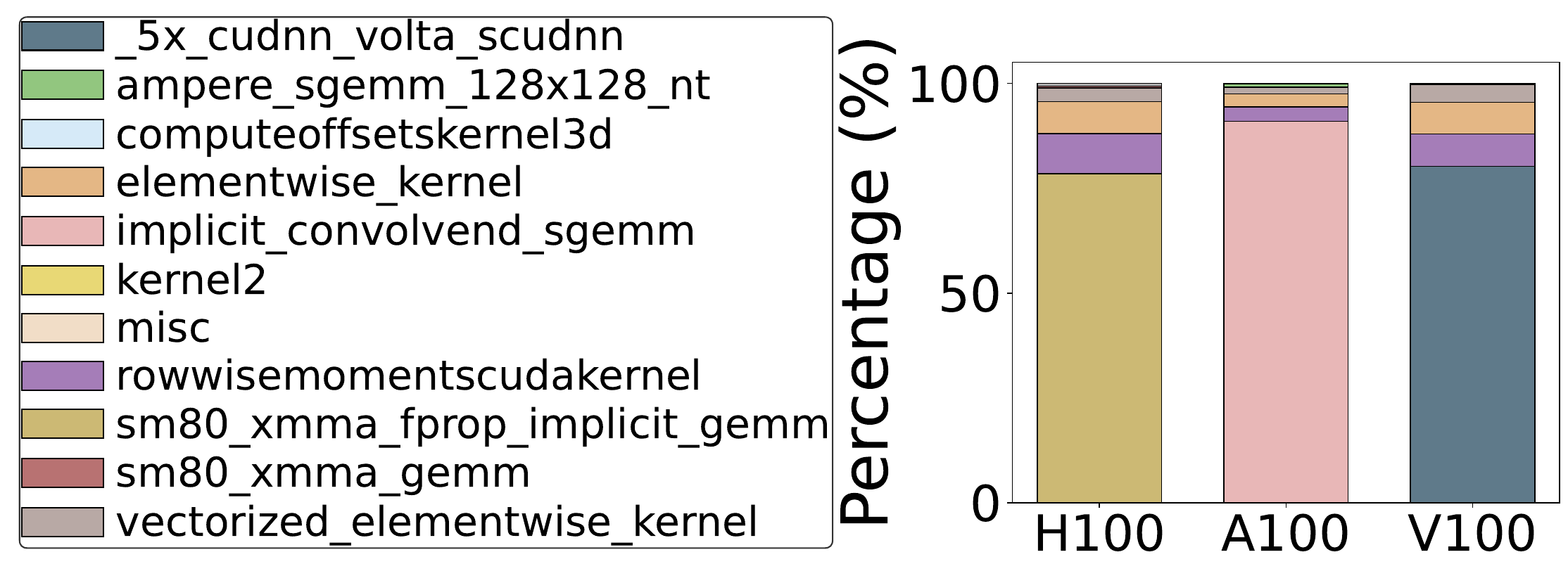}
\vspace{-5pt}
  \caption{Med-DDPM Kernel Mix}
\vspace{-10pt}
  \label{fig:kernel_mix}
\end{figure}

Figure~\ref{fig:kernel_mix} presents the kernel-level composition of Med-DDPM’s reverse-diffusion loop across V100, A100, and H100. Each bar aggregates the proportion of total kernel time spent in major operator classes including cuDNN convolutions, implicit-GEMM kernels, normalization kernels, and a variety of elementwise and layout-conversion kernels. Architecture-specific kernel mix directly shapes hardware utilization patterns, which can lead to imbalanced utilization across compute, memory, and scheduling resources. Consequently, kernel-level characterization is essential for identifying the dominant bottlenecks and guiding targeted optimization strategies. Also, this breakdown highlights how the 3D U-Net structure translates into GPU workloads and makes clear which operations dominate execution on each architecture.

Across all three GPUs, a single cuDNN or implicit-GEMM kernel accounts for the vast majority of runtime. On V100, the
\texttt{\_5x\_cudnn\_volta\_scudnn} convolution kernel consumes nearly \(80\%\) of total execution, with normalization (\texttt{RowwiseMomentsCUDAKernel}) and elementwise kernels contributing only marginally. The skew is even stronger on A100, where \texttt{implicit\_convolveNd\_sgemm} approaches \(90\%\) of runtime. H100 follows the same trend: its primary kernel, \texttt{sm80\_xmma\_fprop\_implicit\_gemm}, accounts for roughly \(78\%\) of observed activity. The remaining kernels' pointwise operations, padding and layout transforms, and small reductions appear only as thin slices in the kernel mix.

\begin{figure}[h]
  \centering
  \includegraphics[width=\linewidth]{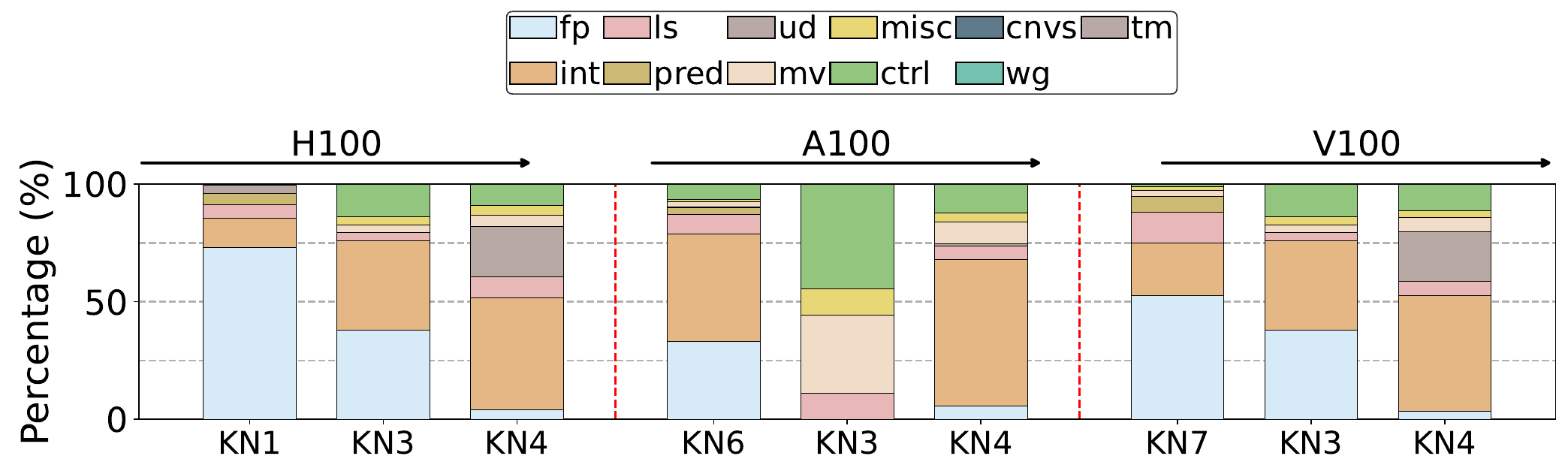}
  \vspace{-4mm}
  \caption{Med-DDPM Instruction Mix}
  \label{fig:imix}
  \vspace{-3mm}
\end{figure}

This distribution reflects the computational structure of the Med-DDPM U-Net, in which 3D convolutions dominate both forward and backward passes and are mapped by PyTorch to cuDNN’s implicit-GEMM backends. Architecturally, the V100’s dominant convolution kernel is mostly constrained by FP32/FP16 FMA throughput rather than by memory bandwidth, leading to a compute-bound regime with limited Tensor Core engagement. A100 and H100, by contrast, route these same convolutional workloads through more advanced MMA pipelines (TF32, BF16, FP16, and Hopper’s enhanced Tensor Cores). Yet despite their higher theoretical throughput, both architectures remain bottlenecked by how efficiently PyTorch and cuDNN manage tile loading, shared-memory reuse, and tensor-layout handling. The persistent dominance of the conv/GEMM pathway across all generations underscores a central result: architectural advances improve throughput but do not alter the fundamental fact that convolution remains the structural bottleneck in Med-DDPM’s training pipeline.

\noindent
\fcolorbox{black}{lpurple}{%
  \parbox{\dimexpr\linewidth-2\fboxsep-2\fboxrule\relax}{%
    \textbf{Takeaway 2:} A single cuDNN convolution/GEMM kernel dominates runtime, making convolution throughput the primary optimization target.
    }%
}

\vspace{-3mm}
\subsection{Instruction-Level Analysis}
The instruction mix in Figure~\ref{fig:imix} provides a complementary view of the underlying bottlenecks from a hardware-focused perspective, showing how each architecture distributes its dynamic instructions within the dominant kernels. Across three architectures, the primary convolution and GEMM kernels contain a notably high proportion of floating-point instructions and comparatively modest load/store activity. On V100, more than half of all dynamic instructions in the main convolution kernel correspond to floating-point operations, with the remainder consisting of integer arithmetic, load/store instructions, and predicate or control flow. This profile reflects a kernel whose behavior is shaped largely by Tensor Core throughput and extensive register reuse rather than by memory-system demands.

A100 exhibits a different distribution, characterized by a larger fraction of integer and control instructions relative to floating-point operations. This pattern aligns with Ampere’s implicit-GEMM design, where MMA pipelines rely on extensive index computation and loop control to coordinate Tensor Core execution. The comparatively small share of load/store instructions indicates that the kernel’s behavior is influenced more by scheduler activity and intra-SM execution than by external memory bandwidth. On H100, the dominant GEMM kernel is even more heavily weighted toward floating-point operations, with over 70\% of its dynamic instructions belonging to FP pipelines. This profile suggests that Hopper’s Tensor Core pathways drive the majority of the kernel’s execution, with memory and control-flow operations contributing minor shares.

The instruction-level characteristics of elementwise kernels differ markedly from those of convolution and GEMM kernels. Elementwise operations contain relatively few floating-point instructions and instead rely heavily on integer arithmetic, uniform datapath operations, register movement, and control flow. This distribution reflects the nature of PyTorch’s fused pointwise kernels, which typically perform simple arithmetic but require substantial address computation, conditional evaluation, and data movement. Such kernels commonly exhibit sensitivity to memory bandwidth, register usage, and warp-level control behavior. Because their arithmetic intensity is low, the overall instruction mix is dominated by operations that support memory access and per-element control rather than direct computation.

Normalization and reduction kernels, such as \texttt{RowwiseMomentsCUDAKernel}, display yet another instruction profile. These kernels rely heavily on load/store instructions, integer arithmetic, and control logic as they repeatedly access feature-map tiles, compute per-channel statistics, and coordinate partial reductions across threads. Their behavior is closely tied to shared-memory bandwidth and warp-synchronous data exchange rather than to floating-point throughput.

Taken together, the instruction-mix data across all architectures reveal distinct execution regimes for convolution, elementwise, and reduction kernels. Convolution and GEMM kernels are dominated by floating-point pipelines and are shaped primarily by Tensor Core execution. Elementwise kernels rely heavily on integer and control operations, reflecting their memory-bound and address-intensive nature. Reduction kernels emphasize load/store and control behavior, consistent with their communication and synchronization patterns. These distinctions illustrate the heterogeneous instruction-level structure of the Med-DDPM model and highlight the diverse hardware resources engaged across different kernel types.

\noindent
\fcolorbox{black}{lpurple}{%
  \parbox{\dimexpr\linewidth-2\fboxsep-2\fboxrule\relax}{%
    \textbf{Takeaway 3:} Convolution, elementwise, and reduction kernels
    stress different instruction paths, requiring multi-faceted rather
    than single-point optimization.}%
}

%% file: 06.optimization.tex
\usetikzlibrary{fit,calc}
\newcommand*{\tikzmk}[1]{\tikz[remember picture,overlay,] \node (#1) {};\ignorespaces}
\newcommand{\boxit}[2]{\tikz[remember picture,overlay]{\node[yshift=3pt,fill=#1,opacity=.25,fit={(A)($(B)+(#2\linewidth,.8\baselineskip)$)}] {};}\ignorespaces}
\colorlet{red}{red!40}
\colorlet{blue}{cyan!60}
\colorlet{blue}{cyan!60}
\colorlet{green}{green!50}
\colorlet{yellow}{yellow!60}

\begin{algorithm}[t]
\footnotesize
\caption{TF32 Tensor-Core Acceleration}
\label{alg:tf32}

\SetKwInOut{Require}{Require}
\SetKwInOut{Ensure}{Ensure}

\Require{3D UNet training loop with convolution- and GEMM-heavy operations}
\Ensure{Use TF32 Tensor Cores on Ampere+ GPUs}

\tikzmk{A}
MatmulBackend.allow\_tf32 = True\;
CuDNNBackend.allow\_tf32 = True\;
\tikzmk{B}
\boxit{red}{0.95}
$M \leftarrow \text{diffusion\_model}$

\ForEach{step $\in$ training\_steps}{
    $(x, y) = \text{select\_batch()}$\;
    \tikzmk{A}
    $M.\text{forward\_tf32\_conv3D}(x, y)$\;
    \tikzmk{B}
    \boxit{blue}{0.89}
    $loss = M.\text{calc\_loss}(x, y)$\;
    \tikzmk{A}
    $M.\text{backpropagate\_tf32\_conv3D\_matmul}(loss)$\;
    \tikzmk{B}
    \boxit{green}{0.89}
    $M.\text{update\_parameters}()$\;
}
\end{algorithm}

\vspace{-2mm}
\section{Optimization Strategies}
In this section, we relate the optimization strategies to the bottlenecks identified in \S\ref{sec:bottleneck}. The kernel and instruction-mix analysis showed that Med-DDPM’s training loop is dominated by convolution and GEMM kernels, with execution shaped largely by Tensor Core throughput and surrounding memory-access patterns. The strategies below therefore target these two factors by enabling high-throughput Tensor Core execution and adopting memory layouts that minimize layout conversions and improve data locality.

\subsection{TF32 Activation for Tensor Core}
Starting from the Ampere architecture, third generation Tensor Cores can be utilized with broader precision support, and we leverage these tensor cores to accelerate high-dimensional tensor operations such as 3D convolution and matrix multiplication, thereby optimizing performance. Algorithm~\ref{alg:tf32} summarizes this TF32 acceleration process. TF32 is an efficient numeric format that combines the wide dynamic range of FP32 with the computation speed of FP16, and it is natively supported by tensor cores. This allows us to achieve significant speedups while maintaining convergence accuracy nearly equivalent to FP32.
The key idea of the algorithm is to explicitly enable TF32 computation in the backend of the PyTorch framework before the training loop begins (\hlword{algored}{the red highlighted}). The \texttt{MatmulBackend.allow\_tf32} setting accelerates matrix multiplication (GEMM) operations used in components such as Linear layers. \texttt{CuDNNBackend.allow\_tf32} setting instructs the cuDNN library to process 3D convolution operations which are most critical for the performance of the 3D U-Net using tensor cores.
These settings directly affect the two most computationally intensive stages of training. First is the forward pass (\hlword{algoblue}{blue highlighted}), where all 3D convolution operations inside the U-Net are accelerated by TF32. Second is the backward pass (\hlword{algogreen}{the green highlighted}), where convolution and matrix multiplication operations required to compute gradients from the loss are efficiently handled by tensor cores.

\begin{algorithm}[t]
\footnotesize
\caption{3D Channels-Last Memory Layout}
\label{alg:channelslast}

\SetKwInOut{Require}{Require}
\SetKwInOut{Ensure}{Ensure}

\Require{Input size $(N, C, H, W, D)$, UNet hyperparameters}
\Ensure{UNet executes with channels-last 3D activations}

\tikzmk{A}
${UNet}$ = create\_UNet(${input\_size}$, ${channels}$, ${blocks}$)\;
${UNet}$ = to\_channels\_last\_3D(${UNet}$)\;
\tikzmk{B}
\boxit{red}{0.95}
${M}$ = GaussianDiffusion(${UNet}$, {size\_of(${H, W, D)}$}, ${timestaps}$)\;

\ForEach{step $\in$ training\_steps}{
    \tikzmk{A}
    $(x, y) = \text{select\_batch()}$\;
    $(x, y) = \text{to\_channels\_last\_3D$(x, y)$}$\;
    \tikzmk{B}
    \boxit{blue}{0.89}
    ${x}_t$ = M.generate\_noisy(${x}$)\; 
    \tikzmk{A}
    $\tilde{x}_t$ = M.predict\_noise(${x}_t$)\;
    \tikzmk{B}
    \boxit{green}{0.89}
    $M.\text{calc\_loss(${x}_t, \tilde{x}_t$)}$\;
    $M.\text{update\_parameters}()$\;
}
\end{algorithm}

\subsection{Channels-Last Memory Layout}
Along with TF32 computation acceleration, we applied a memory layout optimization strategy to maximize the execution efficiency of the 3D U-Net. Modern hardware accelerators such as NVIDIA GPU tensor cores exhibit higher compute throughput and memory bandwidth efficiency when operating on a channels-last memory layout rather than the channels-first layout. Algorithm~\ref{alg:channelslast} provides a detailed overview of how this channels-last memory layout is applied to the training of the 3D diffusion model.

Immediately after creating the base structure of the 3D U-Net model using the \texttt{create\_UNet} function, we convert the model’s memory format to channels-last through the \texttt{to\_channels\_last\_3D function} (\hlword{algored}{the red highlighted}). This transformation modifies the internal structure so that the model’s weights and layers, such as Conv3D, accept and output channels-last tensors instead of channels-first.
Once the main training loop begins (for all steps), the sampled batch’s x (image) and y (mask) remain in the standard format (channels-first). Therefore, immediately before feeding these tensors into the model, we explicitly convert them to a five-dimensional format, where the dimensions follow the order of batch size (N), height (H), width (W), depth (D), and channel count (C), using \texttt{to\_channels\_last\_3D} (\hlword{algoblue}{the blue highlighted}).
When the core training operation \texttt{M.predict\_noise} (\hlword{algogreen}{the green highlighted}) is executed, both the model and the input data are aligned in the channels-last format. As a result, the entire forward and backward pass inside the U-Net executes in a manner that is highly optimized for tensor cores, producing a synergistic effect with TF32 acceleration (Algorithm~\ref{alg:tf32}).

%% file: 07.evaluation.tex
\section{Evaluation}
We evaluate the impact of the proposed optimization strategies on Med-DDPM by comparing four configurations: \textbf{Baseline}, the unmodified Med-DDPM implementation; \textbf{OPT1}, which applies the TF32 tensor-core optimization in isolation; \textbf{OPT2}, which applies the 3D channels-last memory layout optimization in isolation; and \textbf{OPT12}, which integrates both \textbf{OPT1} and \textbf{OPT2} simultaneously. 

\begin{figure}[h]
  \centering
  \includegraphics[width=\linewidth, keepaspectratio]{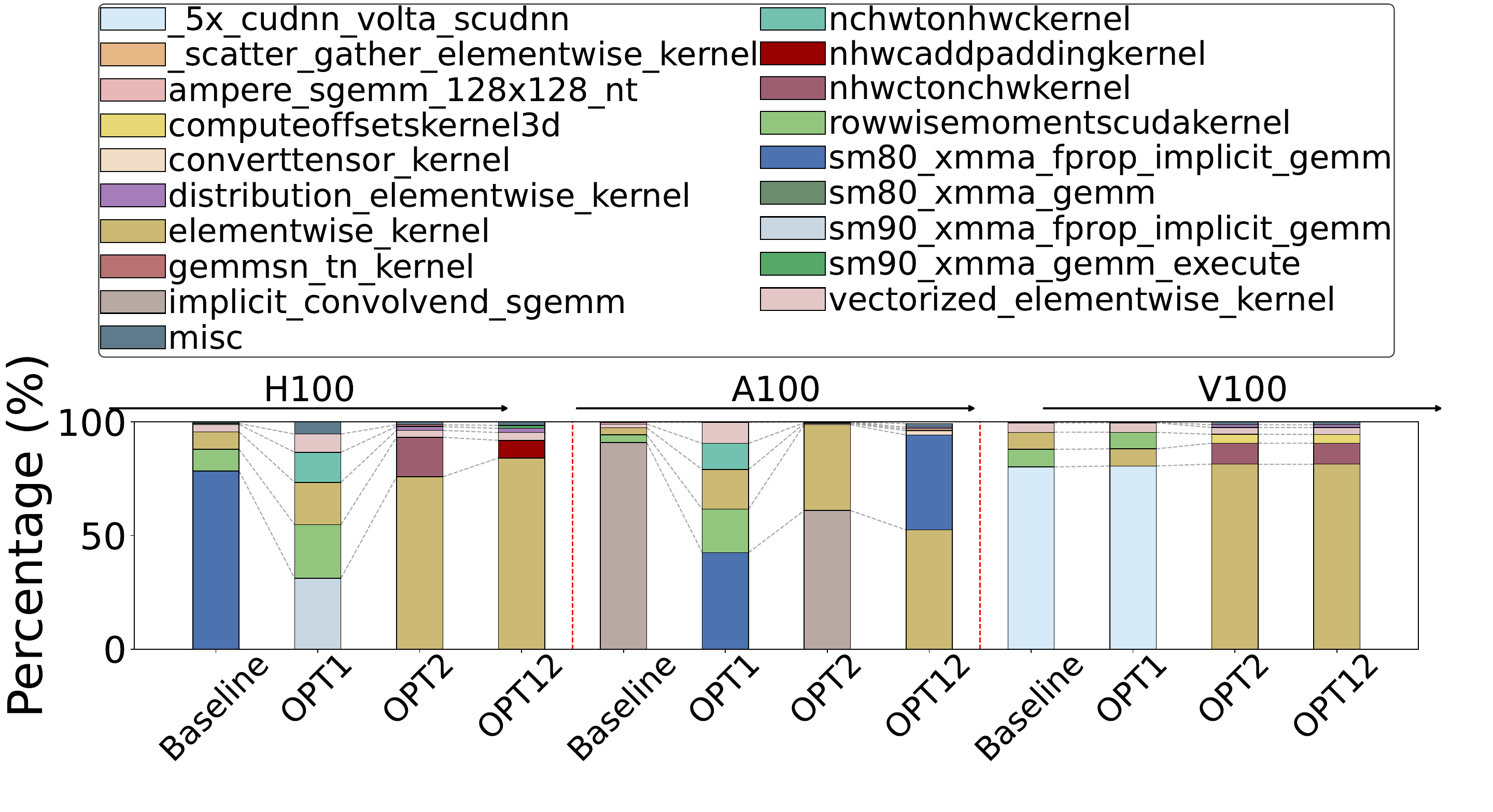}
\vspace{-20pt}
  \caption{Kernel Mix Bar Chart for Optimizations}
\vspace{-2mm}
  \label{fig:kernel_mix_opt}
\end{figure}

\vspace{-2mm}
\subsection{Overall Performance Analysis}
Figure~\ref{fig:kernel_mix_opt} shows that baseline Med-DDPM is dominated by large cuDNN convolution/GEMM kernels across all GPUs \texttt{\_5x\_cudnn\_volta\_scudnn} on V100 (\(\sim\)80\%), \texttt{implicit\_convolveNd\_sgemm} on A100 (\(\sim\)91\%), and \texttt{sm80\_xmma\_fprop\_implicit\_gemm} on H100 (\(\sim\)78\%). Enabling TF32 (OPT1) leaves V100 unchanged but substantially reshapes the Ampere/Hopper architectures: on A100, the conv/GEMM share drops to \(\sim\)43\% with row-wise moments and elementwise kernels rising to \(\sim\)19\% and \(\sim\)17\%, while H100 sees Tensor Core work fall to \(\sim\)31\% as normalization, pointwise ops, and layout transforms collectively exceed 55\%. These shifts confirm that TF32 shortens the dominant conv/GEMM path and exposes bottlenecks previously hidden behind long FP32 kernels. Correspondingly, normalized SM cycles and instructions reduce sharply on A100/H100 (cycles at \(0.18\times\) and \(0.39\times\); instructions at \(0.09\times\) and \(0.18\times\)) while IPC remains near baseline (\(1.02\times\) and \(0.95\times\)), consistent with high-throughput MMA instructions compressing the arithmetic footprint without underutilizing the schedulers.

OPT2 (channels-last), by contrast, collapses the conv/GEMM contribution and rebalances execution almost entirely toward \texttt{elementwise\_kernel} (\(\sim\)73--80\%) plus NHWC padding and transform kernels, fragmenting the 3D U-Net into many small memory-bound kernels. This is reflected in extreme reductions in dynamic work on A100/H100 cycles falling to \(0.01\text{--}0.02\times\) and instructions to \(0.01\times\) with IPC slightly increasing (\(\sim\!1.07\times\)) because these lighter kernels retire efficiently at the SM level despite offering little Tensor Core utilization. OPT12 partially restores Tensor Core usage on A100 (GEMM at \(\sim\)42\%), but V100 and H100 remain dominated by elementwise and layout transforms (>80\%), confirming that channels-last fundamentally shifts the bottleneck away from compute throughput. Microarchitecturally, OPT1 is a true arithmetic compression mechanism that preserves SM efficiency via dense MMA execution, whereas OPT2/OPT12 alter the kernel landscape itself, trading away Tensor Core acceleration for a layout-bound pipeline whose performance will only improve through reducing NHWC\(\leftrightarrow\)NCHW transforms, fusing pointwise ops, and improving data locality.

\begin{figure}[t]
  \centering
  \vspace{2mm}
  \includegraphics[trim=0mm 0mm 0mm 0mm,clip,width=\linewidth, keepaspectratio]{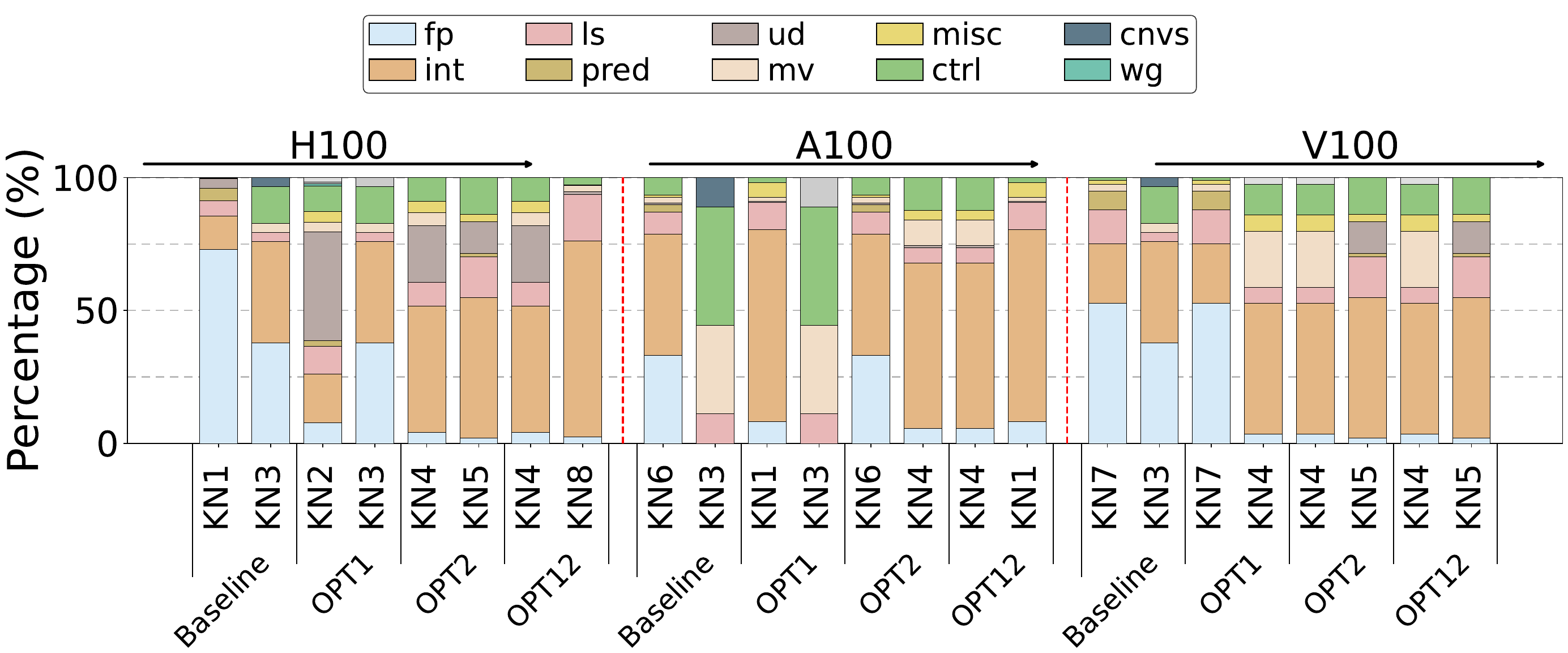}
  \vspace{-5mm}
  \caption{Instruction Mix of Baseline and Optimizations}
  \vspace{-4mm}
  \label{fig:imix_baseline_opt}
\end{figure}

Figure~\ref{fig:imix_baseline_opt} complements this kernel-level view by showing how the dynamic instruction mix evolves under each optimization. On A100, the baseline \texttt{implicit\_convolveNd\_sgemm} kernel is split between integer and floating-point work (Integer \(\approx 45\%\), FP \(\approx 33\%\)), with the remainder spread across load/store (\(\approx 8\%\)), control (\(\approx 7\%\)), and small fractions of movement, predicate, and uniform-datapath instructions. Under OPT1, the corresponding Tensor Core path becomes strongly integer-heavy (Integer \(\approx 72\%\), FP \(\approx 10\%\)), while on H100 the FP share drops from \(\approx 73\%\) to \(\approx 41\%\) and uniform-datapath plus movement/miscellaneous instructions roughly double. This shift reflects how TF32 MMAs are implemented as compact matrix micro-ops that consume integer and uniform-datapath resources rather than a long stream of scalar FP instructions, increasing predicate/control traffic for tile scheduling but shrinking the FP portion of the dynamic mix. In contrast, the channels-last pipeline (OPT2/OPT12) moves the aggregate mix toward the profile of elementwise and layout kernels, which are dominated by integer address arithmetic, control flow, and load/store/movement operations. On Ampere/Hopper, this is visible in the elevated shares of control, predicate, and movement instructions in Figure~\ref{fig:imix_baseline_opt}, indicating that performance is increasingly limited by index generation and tensor-coordinate conversions rather than by raw floating-point throughput. In summary, OPT1 primarily re-encodes arithmetic into MMA-dominated instruction streams, whereas OPT2/OPT12 restructure Med-DDPM into a memory- and layout-bound regime with substantially higher non-arithmetic instruction overhead.

\vspace{-3mm}
\subsection{Microarchitectural Analysis}
\begin{figure}[t]
  \centering
  \includegraphics[width=1\linewidth, keepaspectratio]{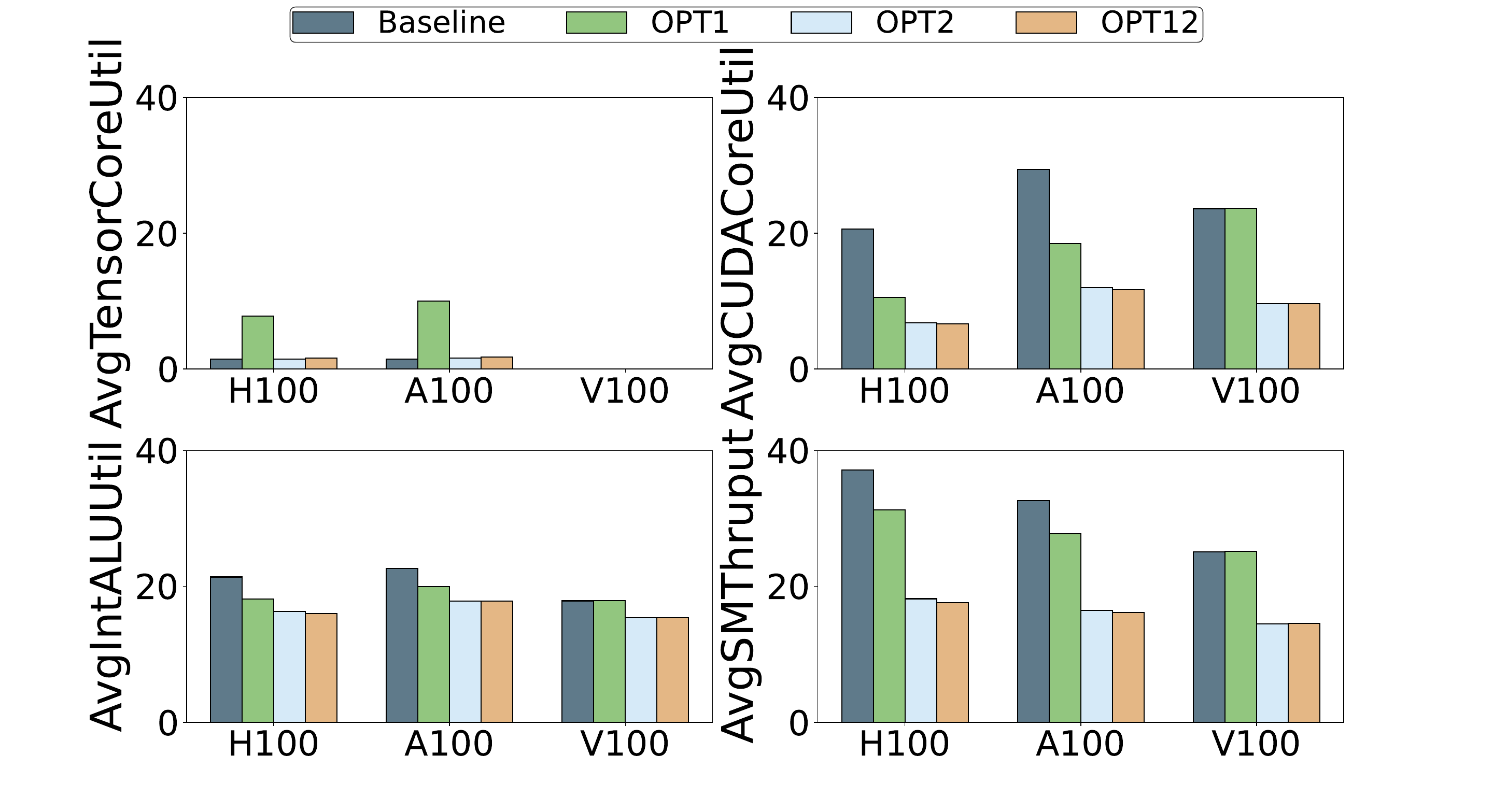}
\vspace{-25pt}
  \caption{Average Tensor/CUDA/Int ALU Core Utilization and SM Throughput}
\vspace{-10pt}
  \label{fig:tensor_util}
\end{figure}

The hardware-usage counters reinforce the earlier kernel-mix and cycle/IPC analysis by making explicit how OPT1 redistributes work between Tensor Cores and the traditional CUDA FP32 pipelines on Ampere and Hopper. On H100 and A100, baseline Med-DDPM barely engages Tensor Cores (\(\text{Avg Tensor Core Util} \approx 1.4\%\)), consistent with the dominance of FP32 cuDNN-style convolutions. Under OPT1, Tensor Core utilization jumps to \(7.81\%\) on H100 and \(9.98\%\) on A100, a \(5\text{--}7\times\) increase, while average CUDA core utilization drops from \(20.60\%\rightarrow 10.53\%\) (H100) and \(29.39\%\rightarrow 18.46\%\) (A100) as shown in Figure~\ref{fig:tensor_util}. This matches the earlier observation that TF32 activation compresses scalar FMA work into dense MMA instructions: the SM warp schedulers feed a smaller number of Tensor Core–resident conv/GEMM instructions that retire far more arithmetic per issue, allowing the FP32 pipelines to idle more without hurting throughput. The modest reduction in average SM throughput (\(37.14\%\rightarrow 31.27\%\) on H100 and \(32.64\%\rightarrow 27.77\%\) on A100) should therefore be read alongside the large drops in cycles and instructions: the SMs are less “busy” in a normalized sense because there is simply less dynamic work to perform per sample, not because they are starved. The integer ALU utilization also decreases slightly (e.g., H100 \(21.38\%\rightarrow 18.11\%\), A100 \(22.66\%\rightarrow 20.00\%\)), indicating that address-generation and loop-control overheads shrink with the shorter, more Tensor Core–dominated conv loops. On V100, Tensor Core utilization remains at zero for all configurations, and both CUDA core and SM throughput metrics stay essentially flat between baseline and OPT1 (\(\text{Avg CUDA Core Util} \approx 23.6\%\), \(\text{Avg SM Thruput} \approx 25.1\%\)), which is consistent with the cycle, instruction, and kernel-mix data.
Volta operates exclusively on the FP32 convolution path, as it lacks TF32-capable Tensor Cores.

The channels-last optimization (OPT2) and its combination with TF32 (OPT12) expose a different architectural regime where the SMs are underutilized even as total cycles and instructions collapse. On H100, OPT2 and OPT12 leave Tensor Core utilization near baseline levels (\(1.47\%\) and \(1.62\%\) versus \(1.44\%\)), while CUDA core utilization falls sharply to \(6.81\%\) and \(6.66\%\) and SM throughput to \(18.18\%\) and \(17.60\%\), approximately half of the baseline. A100 shows the same pattern: Tensor Core utilization only nudges up to \(1.59\%\text{--}1.76\%\), whereas CUDA core utilization drops from \(29.39\%\) to \(12.01\%\text{--}11.69\%\) and SM throughput from \(32.64\%\) to \(16.48\%\text{--}16.16\%\). 
Combined with the kernel-mix data which shows that channels-last shifts execution toward elementwise kernels and NHWC-related padding and transform operations the hardware counters indicate that the workload has entered a highly memory-bound, low-arithmetic-intensity regime.
At the same time, neither CUDA cores nor Tensor Cores are being driven near their compute limits as the SMs spend more time stalled on data movement and kernel launch/tear-down overhead than doing floating-point work. On V100, OPT2/OPT12 drop CUDA core utilization from \(23.63\%\) to \(9.65\%\) and SM throughput from \(25.11\%\) to \(\sim 14.5\%\), with Tensor Core utilization remaining at zero; this aligns with the earlier observation of much fewer total instructions and cycles but a higher relative IPC, implying that when warps do execute, they use the FP32 pipelines more efficiently, yet the overall SM residency is lower due to fragmentation into many small elementwise/layout kernels. Taken together, these hardware-usage trends show that OPT1 achieves genuine compute-path compression by shifting work from FP32 cores to high-throughput Tensor Cores on Ampere/Hopper, whereas the current channels-last implementation (OPT2/OPT12) reduces overall work but fails to keep either compute engine saturated, making kernel fusion and more aggressive Tensor Core–aware tiling essential for turning the observed cycle reductions into sustained, architecture-scaled speedups.

\begin{figure}[t]
  \centering
  \includegraphics[width=1\linewidth, keepaspectratio]{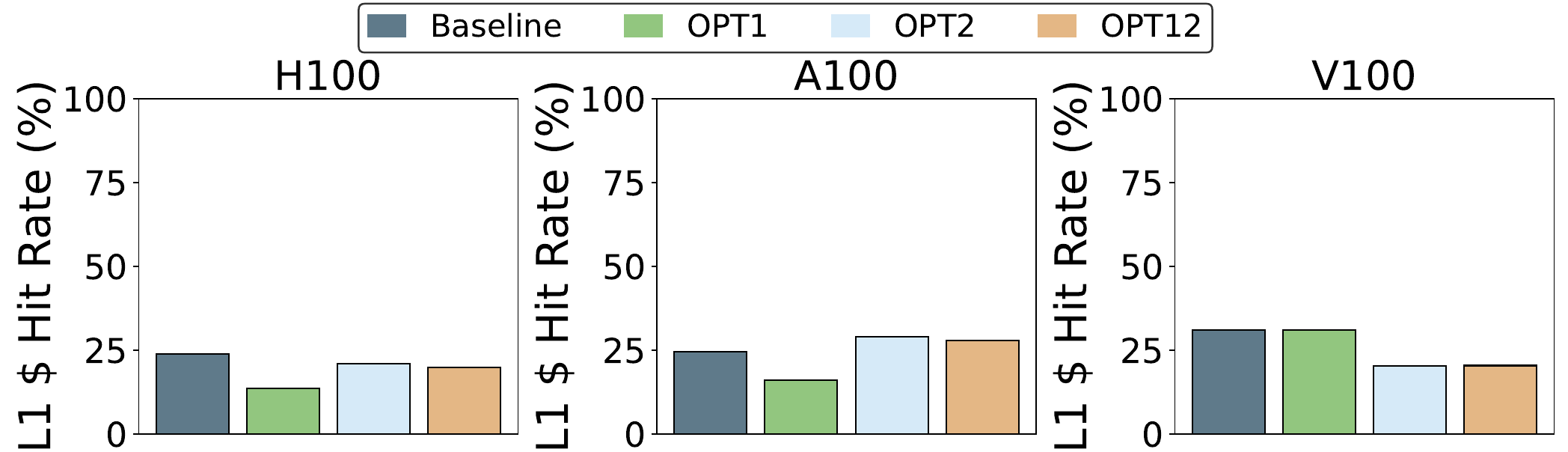}

  \includegraphics[width=1\linewidth, keepaspectratio]{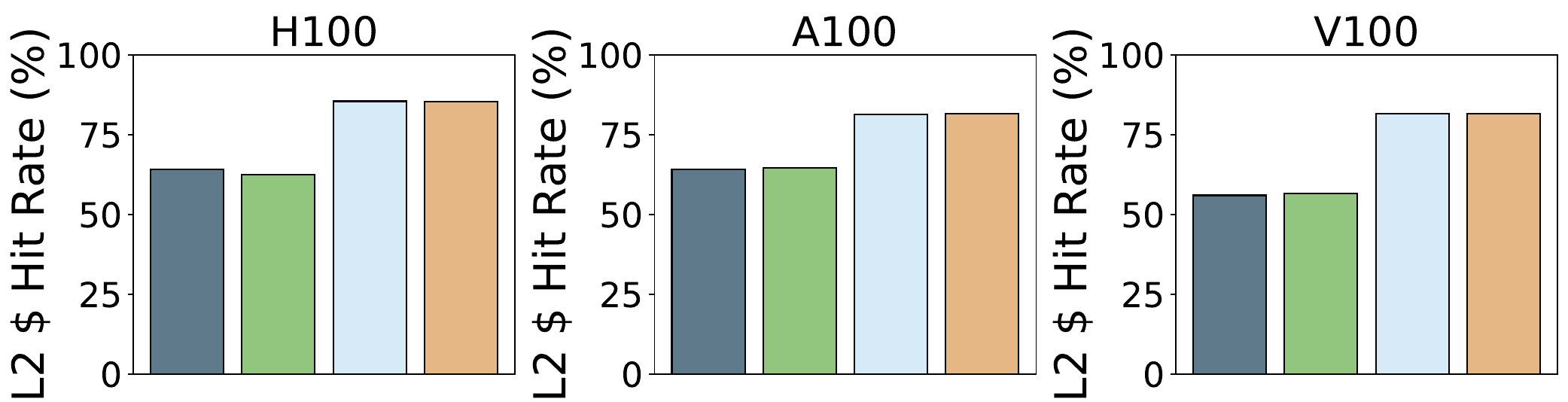}

  \includegraphics[width=1\linewidth, keepaspectratio]{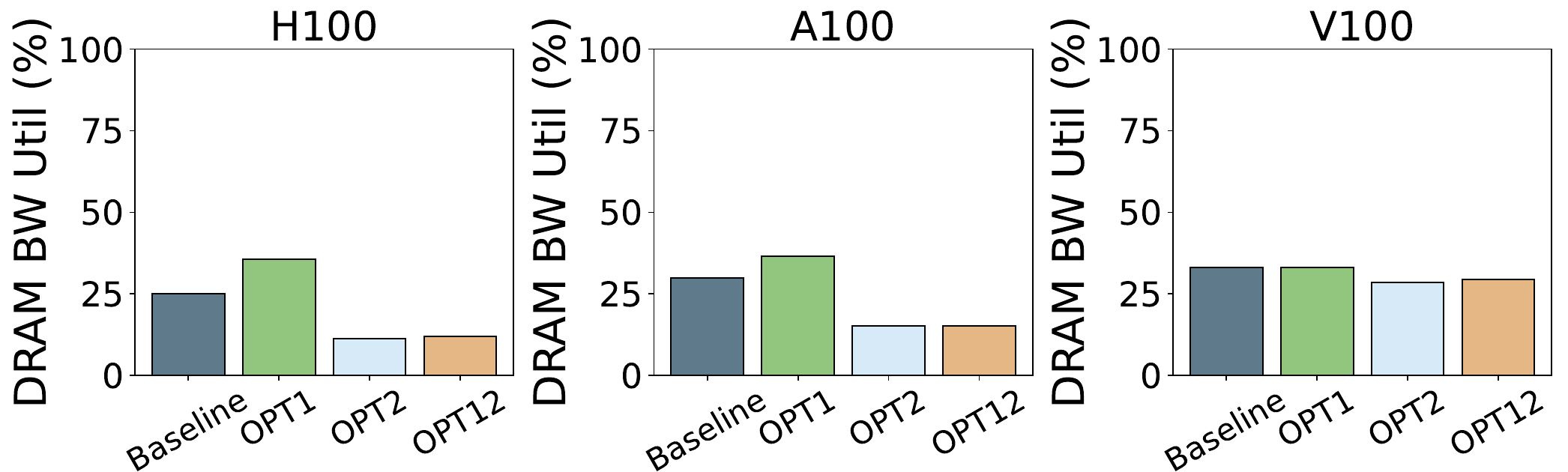}
\vspace{-4mm}
  \caption{L1/L2  Hit Rate, DRAM BW Utilization}
\vspace{-5mm}
  \label{fig:cache_hit}
\end{figure}

\vspace{-3mm}
\subsection{Memory System Analysis}
The cache and DRAM statistics show that OPT1 fundamentally changes how Med-DDPM uses the memory hierarchy on Ampere and Hopper, in a way that is consistent with the Tensor Core–centric execution pattern identified earlier. On H100 and A100, OPT1 reduces the L1 cache hit rate from \(23.86\%\rightarrow 13.51\%\) and \(24.50\%\rightarrow 16.15\%\), respectively, while leaving the L2 cache hit rate essentially unchanged relative to baseline (\(64.10\%\rightarrow 62.44\%\) on H100 and \(64.21\%\rightarrow 64.63\%\) on A100), as shown in Figure~\ref{fig:cache_hit}. At the same time, DRAM bandwidth utilization increases sharply, from \(25.00\%\rightarrow 35.66\%\) on H100 and \(29.95\%\rightarrow 36.55\%\) on A100, with V100 showing flat L1/L2 behavior and essentially unchanged DRAM utilization. Microarchitecturally, this pattern suggests that TF32-enabled Tensor Core convolutions are issuing larger, more contiguous global-memory tiles that stream through the SMs with less fine-grained temporal locality at L1. The per-tile reuse window is narrower, so fewer accesses are captured in the small L1, but the accesses are still regular enough to be serviced efficiently by the L2. The higher DRAM bandwidth utilization reflects the fact that, per unit of SM active time, the Tensor Core path pulls data more aggressively from memory to keep the MMA pipelines fed, even though the total number of cycles and instructions per sample decreases. In other words, OPT1 trades some L1 locality for higher sustained memory throughput into Tensor Cores, aligning with the earlier observation that conv/GEMM kernels shrink in time while Tensor Core utilization rises and CUDA core/INT ALU activity drops.

In contrast, the channels-last optimization (OPT2) and its combination with TF32 (OPT12) push the workload into a regime where the L2 cache absorbs most of the traffic and DRAM is significantly underutilized, reinforcing the view that these configurations are dominated by low-intensity elementwise and layout kernels rather than dense convolutions. Across all three architectures, the L2 hit rate jumps dramatically under OPT2/OPT12, from \(64.10\%\rightarrow 85.47\%\) on H100, \(64.21\%\rightarrow 81.40\%\) on A100, and \(56.03\%\rightarrow 81.57\%\) on V100 (with nearly identical values for OPT12), while DRAM bandwidth utilization collapses to \(11.32\%\text{--}11.87\%\) on H100 and \(15.14\%\text{--}15.15\%\) on A100 and only slightly decreases on V100 (\(33.07\%\rightarrow 28.39\%\)). L1 behavior becomes more architecture- and layout-sensitive: on A100, the L1 hit rate \emph{increases} to \(29.03\%\) (OPT2) and \(27.84\%\) (OPT12), consistent with channels-last improving spatial locality for many pointwise and padding kernels, whereas on H100 and V100, the L1 hit rate either recovers only partially from OPT1 or drops below baseline. Taken together with the earlier kernel-mix and SM utilization results, these memory statistics indicate that OPT2/OPT12 dramatically reduce the amount of data that must be fetched from DRAM and concentrate traffic in L2, but they do so by replacing large, high-flop Tensor Core convolutions with numerous small, memory-bound kernels whose working sets fit comfortably in cache. The SMs thus spend more time stalled on control and launch overhead with neither Tensor Cores nor CUDA cores saturated, and the memory system is over-provisioned relative to the reduced arithmetic intensity. This explains why OPT2/OPT12 achieve large reductions in cycles and instructions without translating into proportionate, architecture-scaled speedups. The memory hierarchy is no longer the bottleneck, but the compute engines are also not being driven at their design limits.

\begin{figure}[t]
  \centering
  \includegraphics[trim=0mm 0mm 0mm 45mm,clip,width=\linewidth, keepaspectratio]{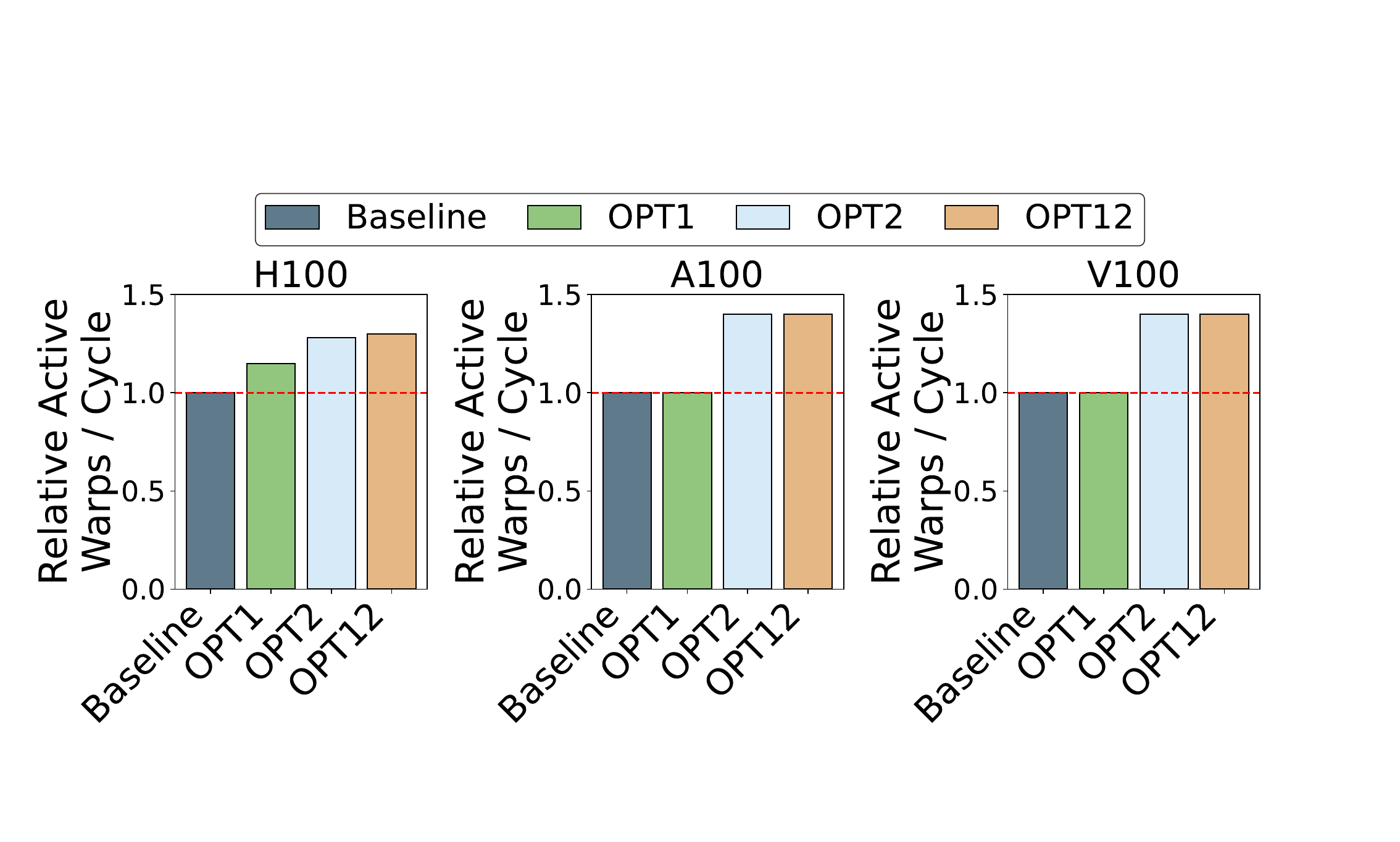}
\vspace{-12mm}
  \caption{Relative Active Warps/Cycle}
\vspace{-4mm}
  \label{fig:warp_per_cycle}
\end{figure}

\vspace{-3mm}
\subsection{Warp Efficiency Analysis}
The warp-level behavior further clarifies the execution regimes created by the two optimizations and aligns closely with the earlier kernel, memory, and SM-utilization findings. Figure~\ref{fig:warp_per_cycle} presents relative active warps per cycle. OPT1 shows only modest changes in active warps per cycle H100 increases slightly to \(1.15\times\) baseline while A100 and V100 remain at \(1.00\times\) because TF32-accelerated convolutions still launch large, well-tiled kernels whose occupancy and scheduling structure broadly match the baseline FP32 path. The main effect of OPT1 is reducing dynamic instructions and cycles, not altering warp residency. In contrast, OPT2 and OPT12 substantially increase relative active warps on three GPUs, reaching \(\!1.30\times\) on H100 and \(1.40\times\) on A100/V100. This reflects the fragmentation of 3D U-Net execution into many smaller elementwise, padding, and layout kernels whose working sets fit comfortably in L2 and often in L1, enabling high resident-warp counts even as arithmetic intensity collapses. These kernels are lightweight enough that the scheduler can keep more warps in flight, but as shown by the drastic drops in SM throughput, Tensor Core utilization, and DRAM bandwidth, they do not generate sufficient compute or memory pressure to saturate the hardware. Thus, high active-warp density under OPT2/OPT12 does not translate into performance as
Warp occupancy is high, but execution efficiency remains low. This mirrors the earlier observation that the channels-last path shifts the workload from dense, compute-bound convs into a high-occupancy yet low-efficiency regime dominated by memory-bound micro-kernels.

\begin{figure}[t]
  \centering
  \includegraphics[width=1\linewidth, keepaspectratio]{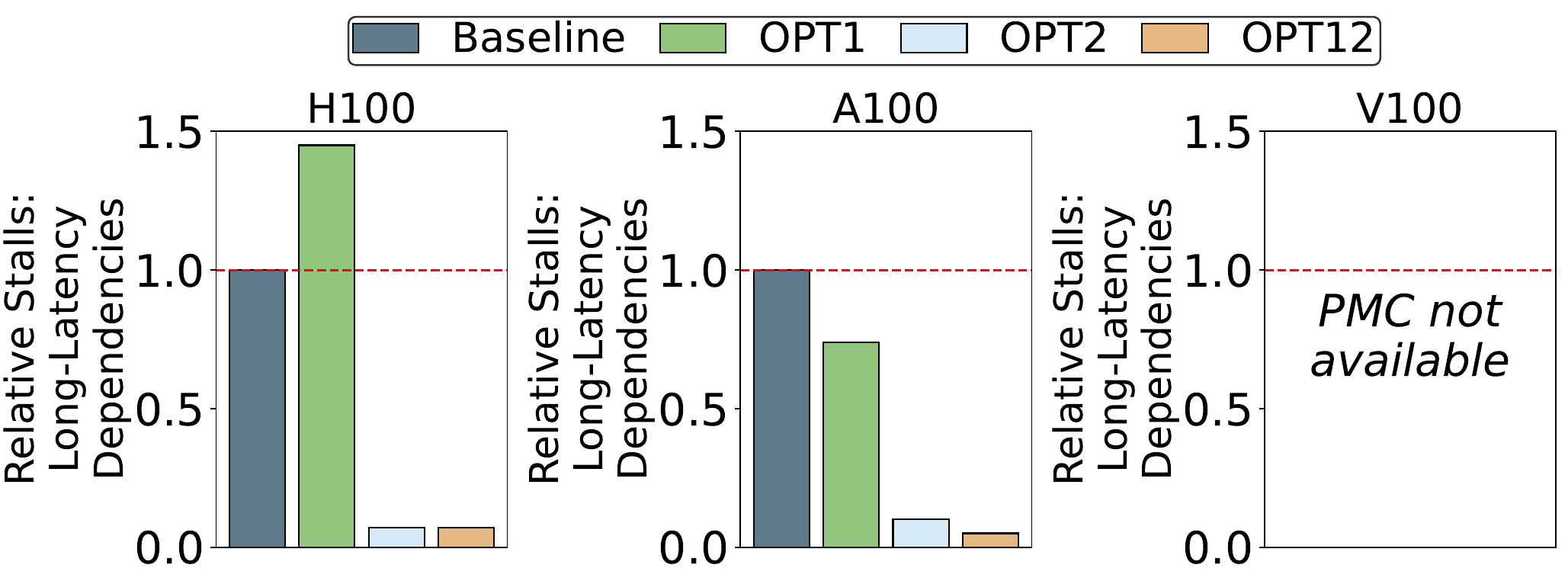}

  \includegraphics[width=1\linewidth, keepaspectratio]{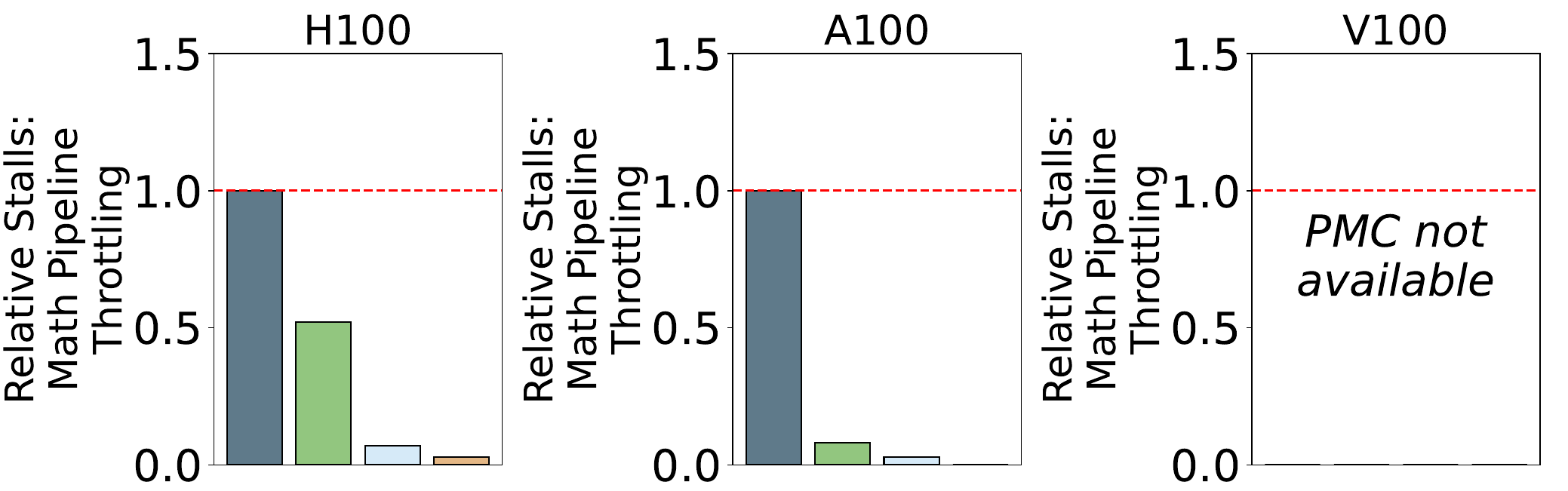}

  \includegraphics[width=1\linewidth, keepaspectratio]{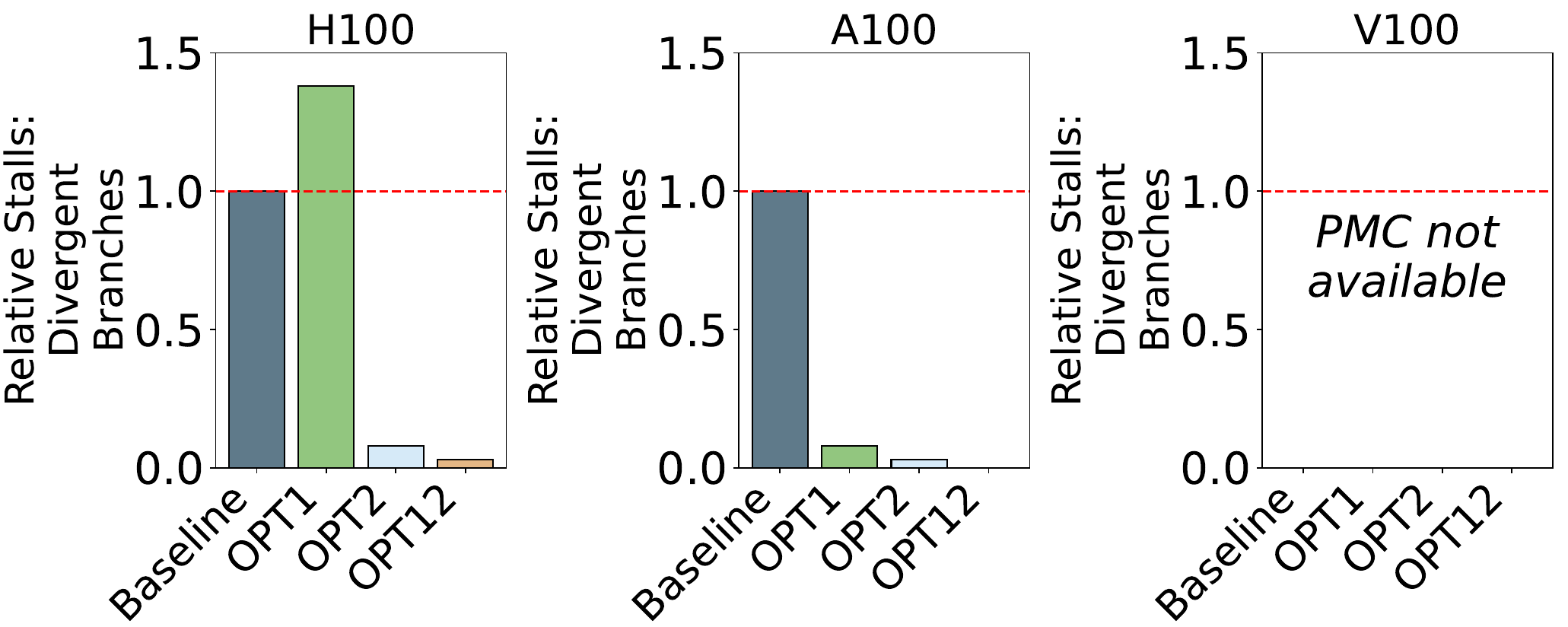}
\vspace{-4mm}
  \caption{Relative Stall Breakdown}
\vspace{-6mm}
  \label{fig:stall}
\end{figure}

\vspace{-3mm}
\subsection{Scheduling Efficiency Analysis}
\label{subsec:schedule_analysis}
The scheduler-level stall breakdown clarifies why OPT1 shifts Med-DDPM into a Tensor-Core–dominated execution regime on Ampere and Hopper and aligns with the earlier kernel, memory, and warp analyses. Figure~\ref{fig:stall} shows relative stalls due to dependencies, throttling, and divergent branches. On H100, OPT1 increases long-latency-dependency stalls to \(1.45\times\) baseline and divergent-branch stalls to \(1.38\times\), while sharply reducing math-pipeline throttling to \(0.52\times\). This pattern reflects Tensor Core–enabled convolutions that issue large, deep MMA tiles whose operand availability often depends on multi-stage global-memory pipelines (as previously evidenced by lower L1 hit rates, stable L2 behavior, and significantly higher DRAM bandwidth). The drop in math-throttling stalls indicates that the FP32 pipelines are no longer the bottleneck Instead, the SMs increasingly wait on long dependency chains for Tensor Core operands, spanning global-memory loads, shared-memory staging, and tile setup. A100, however, shows a smoother schedule under OPT1. All stall categories decrease (long-latency \(0.74\times\), math throttling \(0.08\times\), divergent \(0.08\times\)), consistent with the earlier findings that TF32 greatly reduces cycles and instructions while keeping IPC near or above baseline.

In contrast, OPT2 and OPT12 push the model into a near “stall-free” regime on both H100 and A100, with all stall categories collapsing to \(0.03\text{--}0.10\times\) baseline. When viewed alongside the extremely high L2 hit rates, minimal DRAM bandwidth usage, and elevated active-warp density, this reveals that the channels-last path decomposes the 3D U-Net into many short, cache-resident elementwise/layout kernels with minimal dependency depth and little opportunity for either math-pipeline pressure or long memory-latency exposure. However, as earlier SM-throughput and Tensor/Core CUDA utilization results showed, this state is not performance optimal: although warps rarely stall, they also perform very little useful computation per issue, fail to engage Tensor Cores meaningfully, and operate at low arithmetic intensity. Thus, OPT2/OPT12 eliminate classic stall sources not because the pipeline is maximally utilized, but because the workload has been transformed into a high-occupancy, low-productivity sequence of micro-kernels explaining why dramatic reductions in cycles and instructions do not translate into architecture-scaled speedups. On V100, all three stall classes remain effectively zero across configurations, confirming that neither TF32 nor channels-last materially alters its FP32-centric scheduling behavior.

%% file: 08.related_work.tex
\vspace{-4mm}
\section{Related Work}

\textbf{Medical Imaging using Machine Learning:}
Machine learning has reshaped medical imaging by enabling automated diagnosis and prognosis prediction. Recent advances in deep learning architectures have significantly improved performance in tumor detection, organ segmentation, and disease classification~\cite{laci2025deep, kumar2025advances, galic2023machine, lawrence2025ai, li2023medical}. The models trained on multimodal datasets now support precision oncology by integrating radiology, pathology, and genomics data. Also, regulatory bodies such as the FDA have begun approving AI-based imaging tools, underscoring their clinical relevance~\cite{waqas2024multimodal, xu2025generalizable, zhao2023leuda, zhang2024leveraging, samala2024ai}.

\noindent\textbf{Denoising Diffusion Probabilistic Models:}
Denoising Diffusion Probabilistic Models (DDPMs) have emerged as powerful generative frameworks for medical image synthesis and reconstruction~\cite{dorjsembe2024conditional,zhou2024simple,ho2020ddpm}. Unlike GANs, DDPMs offer stable training and controllable sampling, making them suitable for high-fidelity 3D medical imaging. Recent work such as Medical Diffusion demonstrates DDPMs applied to volumetric MRI generation, achieving realistic anatomical structures~\cite{guo2024reconformer, xu2024stage, liu2025plug}. However, the computationally heavy workload of these models demands the need for optimized GPU kernels and efficient architectural approaches to make training and inference practical at scale.~\cite{liu2025plug, Park_2025_ICCV, liu2024evolving, jiang2025fast, luo2024measurement}.

\noindent\textbf{Synthetic Image Generation:}
Synthetic data generation addresses key limitations in medical imaging, including data scarcity, privacy constraints, and annotation cost~\cite{irmakci2022multicontrast, hsu2022synthetic, sizikova2024synthetic}. GAN-based methods have been widely used to generate synthetic radiographs, CT scans, and MRIs~\cite{pozzi2024generating, luschi2025design, gaggion2023improving}. More recently, diffusion models have been employed to produce high-resolution synthetic 3D images with anatomical labels~\cite{silva2025designing, wang2025self}. Generating such datasets requires significant load on GPU clusters, requiring efficient design of data pipelines, mixed-precision execution, and distributed data generation workflows to maintain throughput.~\cite{nazir2025diffusion, peng2022knowledge}.

\noindent\textbf{Performance Analysis of Diffusion Models:}
Diffusion models are computationally intensive, requiring GPU-level optimization. Recent work explores acceleration techniques such as FP8 quantization and TensorRT integration to reduce latency and cost on Hopper GPUs~\cite{zhao2025radiomics, ekelund2025boosting}. However, there is no in-depth profiling of diffusion models. ~\cite{liu2025mambadiff, chen2024lowbitwidth, dombrowski2024tradeoffs}. Profiling tools like Nsight Compute provide performance analysis with PMCs (Performance Monitoring Counters), guiding kernel fusion and layout transformations. Studies also emphasize the role of memory bandwidth and launch overhead in shaping performance bottlenecks~\cite{kong2024cambricond, wimbauer2024cache}.

%% file: 09.conclusion.tex
\section{Conclusion}
This work provides a comprehensive, system-level analysis of Med-DDPM across three NVIDIA GPU architectures and demonstrates that 3D diffusion workloads are dominated by cuDNN convolution and implicit-GEMM kernels whose inefficiencies stem from memory-access behavior, tensor-layout conversions, and limited Tensor Core engagement. Using detailed Nsight Compute profiling, we show that enabling TF32 Tensor Core paths and adopting a 3D channels-last layout materially reshapes the microarchitectural execution profile. TF32 consistently compresses arithmetic work on A100 and H100, reducing SM cycles by up to 5$\times$ while preserving IPC and memory balance, whereas the channels-last layout exposes a new regime dominated by elementwise and layout kernels that achieve high cache locality but underutilize compute pipelines. Together, these findings highlight that effective acceleration of large 3D diffusion models requires aligning precision modes, memory formats, and kernel behavior with underlying GPU microarchitecture, and they establish TF32 as a robust optimization baseline.